\documentclass[5p]{elsarticle}

\usepackage{hyperref}

\nonstopmode

\usepackage[american]{babel}
\usepackage{bbm}
\usepackage{amsmath}
\usepackage{amssymb}
\usepackage{graphicx}
\usepackage{color}
\usepackage{breqn}

\usepackage{url}
\usepackage{xcolor}
\definecolor{newcolor}{rgb}{.8,.349,.1}

\journal{Signal Processing: Image Communication}
\begin{document}

\begin{frontmatter}

\title{Segmentation of turbulent computational fluid dynamics simulations with unsupervised ensemble learning}
\author[1,2]{Maarja Bussov%
    \corref{cor1}%
    }
    \ead{bussovm@ut.ee}
    \address[1]{Tartu Observatory, University of Tartu, Observatooriumi 1, 61602-zip, T\~oravere, Estonia}
    \address[2]{Department of Physics, University of Helsinki, P.O. Box 64, FI-00014, Helsinki, Finland}     
    \cortext[cor1]{Corresponding author}

\author[3,4]{Joonas N\"attil\"a%
    }
    \ead{jan2174@columbia.edu}
    \address[3]{Physics Department and Columbia Astrophysics Laboratory, Columbia University, 538 West 120th Street, New York, NY 10027, USA}
    \address[4]{Center for Computational Astrophysics, Flatiron Institute, 162 Fifth Avenue, New York, NY 10010, USA}

\begin{abstract}
    Computer vision and machine learning tools offer an exciting new way for automatically analyzing and categorizing information from complex computer simulations.
    Here we design an ensemble machine learning framework that can independently and robustly categorize and dissect simulation data output contents of turbulent flow patterns into distinct structure catalogues.
    The segmentation is performed using an unsupervised clustering algorithm, which segments physical structures by grouping together similar pixels in simulation images.
    The accuracy and robustness of the resulting segment region boundaries are enhanced by combining information from multiple simultaneously-evaluated clustering operations.
    The stacking of object segmentation evaluations is performed using image mask combination operations.
    This statistically-combined ensemble (SCE) of different cluster masks allows us to construct cluster reliability metrics for each pixel and for the associated segments without any prior user input.
    By comparing the similarity of different cluster occurrences in the ensemble, we can also assess the optimal number of clusters needed to describe the data.
    Furthermore, by relying on ensemble-averaged spatial segment region boundaries, the SCE method enables reconstruction of more accurate and robust region of interest (ROI) boundaries for the different image data clusters.
    We apply the SCE algorithm to 2-dimensional simulation data snapshots of magnetically-dominated fully-kinetic turbulent plasma flows where accurate ROI boundaries are needed for geometrical measurements of intermittent flow structures known as current sheets.
\end{abstract}

\begin{keyword}
plasma ---
turbulence ---
clustering analysis ---
image segmentation ---
unsupervised machine learning\\
\texttt{elsarticle.cls}\sep \LaTeX\sep Elsevier \sep template
\MSC[2010] 00-01\sep  99-00
\end{keyword}

\end{frontmatter}

\section{Introduction}

Turbulent, seemingly chaotic flows, are known to 
create self-similar structures and flow patterns on multiple scales. 
Especially interesting are any discrete long-lived structures that appear close to the dissipation scales, originating from the intermittency of the turbulence.
In magnetohydrodynamical (MHD) turbulence these intermittent structures manifest themselves as the so-called current sheets that are regions of intense electric current flowing in thin two-dimensional sheet-like configurations \citep{Biskamp_2003}.
Analyzing the geometrical shapes and sizes of these structures can help us better quantify the role of intermittency in turbulent fluids and plasmas.

In the past, different statistical methods for automating the detection of such structures have been conceived but the algorithms are known to be computationally very demanding \citep[e.g.][]{Uritsky2010, Zhdankin2013, Azizabadi2020}.  
Machine learning methods and computer vision algorithms offer a new promising avenue for constructing such detection frameworks since they are fast to evaluate and highly optimized \citep[e.g.,][]{2020Dupuis, Hu2020,2021Sisti}.
Here, we present an ensemble unsupervised machine learning algorithm for image structure detection and automated segmentation that is specifically tailored for structure detection from computer simulation outputs. 

In general, computer vision algorithms and the related machine learning tools offer an interesting new way of studying physical systems. 
These algorithms enable an easy-to-use and efficient automation of visual segmentation of computer simulation outputs.
Image segmentation is an actively studied problem in the field of computer vision where the aim is to label each pixel in an image into a distinct category.
Typically, supervised learning models are used to perform the pixel category assignment --- in this case, large training sets containing labels for each pixel needs to be compiled a priori.
The state-of-the-art models for real-world images are prominently deep learning algorithms \citep{he2017mask, choy20194d, Valada_2019} and they are rapidly evolving and improving \citep[e.g.,][]{Landrieu2018PointCloud, wang2017nonlocal, Zhu2019AsymmetricNN}.
However, the large computational cost and requirement for a vast pixel-level labeled pool of data make some of the best performing supervised models not suitable for practical analysis.
Additionally, obtaining a desired accuracy for deep learning models is known to be notoriously hard
\citep{HURTADO2001, Shrestha2019}.

A major class of machine learning methods tackling data with no ground truth is the unsupervised learning method.
Unsupervised learning algorithms are prominently used for cluster detection, dimensionality reduction of high dimensional data, or aiming at representing the data with a few prototypes.
Unsupervised algorithms, which can be applied to data with no knowledge of the ground truth or even of the amount of possible clusters, offer an easy-to-use alternative for segmentation of images and videos \citep{gansbeke2020scan, Ventura2019RVOSER}.
A major drawback of unsupervised machine learning methods is that the resulting image segments and their associated clusters tend to be varying from one evaluation of the algorithm to another.
This can be highly unwanted especially in applications where robust cluster classification, segmentation, and pixel-level image dissection are needed.
The validity of unsupervised clustering algorithms depends on the algorithm and prior knowledge about the data; 
this estimation is usually done with external or internal metrics \citep{halkidi2001clustering, Rini_2018}.

A common cure for increasing the "signal strength" of individual analysis results is to combine multiple evaluations together, hence increasing the signal-to-noise ratio.
These algorithms are typically known as ensemble machine learning methods. 
Ensemble frameworks have been developed to increase the robustness and stability of unsupervised clustering algorithms in many previous studies \citep{Fred2002, zhang2014unsupervised, topchy2004mixture}.
Graph partitioning models to create cluster ensembles show promising results in \citep{GraphPartitioning1, strehl2002cluster}. 
Multimodal hypergraph methods are developed, for example, in \citep{YuIEEE, YuIEEE2}.  
In \cite{jiang2004som} an unsupervised ensemble framework is developed for Self-organizing Map clustering results, similarly to this work.

In this paper we construct a new ensemble framework that combines independent realizations of multiple clustering algorithm results, with the aim to provide reliable and robust segmentation regions obtained via unsupervised learning methods.
The ensemble framework is devised to increase stability and robustness of the clusters detected by an unsupervised learning algorithm from image data. 
We use the Self-Organizing Map (SOM) algorithm \citep[also known as Kohonen's map;][]{kohonenSOM, kohonen2013essentials} as our base clustering algorithm, but labels obtained from other clustering algorithms may also be used.%
\footnote{
The SOM is chosen for its topology preservation quality, ability of obtaining fine structures from image data, and conceptual simplicity.
The SOM algorithm has been used and further developed in thousands of scientific research papers and used in many fields of science including medical sciences, financial sciences, and speech analysis \citep[see e.g.,][]{kohonen2013essentials}.
}
We show that the resulting region of interest (ROI) of the objects from the combined ensemble algorithm are more easily interpretable and geometrically more stable against fluctuations.
This makes the proposed method ideal for automating the segmentation procedure of computer simulations. 
We, however, note that the new ensemble method may also be used for many other similar image segmentation tasks with no ground truth information, where high accuracy of the ROIs are needed like, for example, in medicine, remote sensing, or biology where images need to be dissected and segmented into different structures automatically \citep{2020Qiu, 2020Xiangchun}


\section{Data}
\label{SEC:data}

Our main aim is to dissect the simulation data snapshot pixels into distinct physical structures based on the similarity of their feature vectors.
We are, in particular, interested in automating the grouping of pixels into 
different physical categories whose geometrical size distributions we, in the end, want to measure.

We apply our image segmentation algorithms to fully-kinetic particle-in-cell simulations of decaying relativistic fully-kinetic turbulence \citep[see e.g.,][]{Zhdankin2017, Comisso2018, Nattila2019}. 
Here we omit most of the data interpretation and physics discussion and focus instead on the results and performance of the image segmentation algorithms. 
We refer the reader to \cite{Nattila2019} for a presentation and discussion of the various technical simulation parameters.
We also note that no ground truth data exists in our case.
The situation is common for many non-linear physics-motivated problems that have no a-priori known solution;
instead, both the new algorithm and the system are studied simultaneously.

The analyzed input data is complex, sophisticated and the features are highly correlated.  
It consists of large 2-dimensional rectangular images where each pixel carries multiple features (i.e., a feature vector). 
Generalization to 3-dimensional data input is straightforward; 
here we focus on $2$-dimensional case only to simplify the visualization of the results.
We describe each pixel with a set of restricted invariant physical features obtained from the initial data because image analysis features need to be as invariant as possible \citep{kohonen2013essentials}. 

One simulation image snapshot consists of roughly $1$ million pixels with each pixel storing multiple physical quantities.
We perform the clustering by using $8$ consecutive snapshots from the simulations, with equal time steps between the samplings (corresponding to roughly one eddy turnover time in physical units).
This increases the total number of data points to roughly $8$ million pixels. 
For our segmentation analysis we select a $3-$dimensional data vector 
$X_{k}=(B_\perp, J_{\parallel},  [\vec{J}_\parallel \cdot\vec{E}])$.
These characteristics were chosen by relying on domain knowledge and on the fact that
they capture most of the variability in the data.

Physically we expect that the perpendicular magnetic field averages out, $\langle \delta B_\perp \rangle = 0$, but has a finite root-mean-square value,
 $\sqrt{\langle \delta B_\perp^2 \rangle}/B_0 \approx 1$.
The quantity is, to a first approximation, seen to follow a normal distribution.
The parallel component of the current, 
$J_\parallel$, 
should be small except for a few highly localized regions;
its distribution is therefore expected to strongly deviate from a Gaussian normal distribution.
In our analysis we normalize the value of $J_\parallel$ to the theoretical maximum that a uniform charged particle background can support, yielding $\mathrm{max}(|J_\parallel|) \approx 1$.
The quantity is seen to strongly deviate from a normal distribution due to heavy tails.
The third feature, $\vec{J}\cdot\vec{E}$, is a measure of an energy transport from the electromagnetic fields to the plasma;
for a volumetric dissipation we would expect it to be a constant---in reality its value is highly variable,  reminiscent of the intermittency of the turbulence.
In our segmentation analysis we normalize its value with its root-mean-square value.
The quantity deviates from a normal distribution due to over-pronounced tails.
All of the described features have non-trivial mutual correlations.

\begin{figure*}
   \centering
      \centering\includegraphics[width=6.2cm]{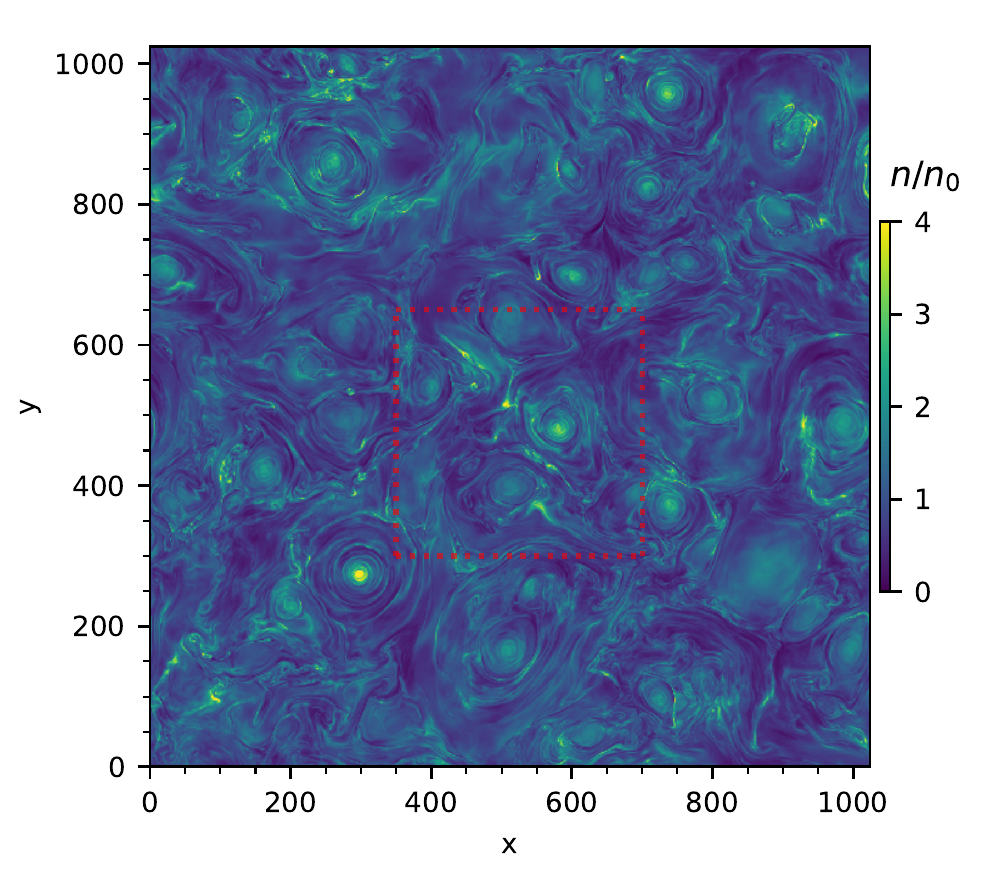}
    \centering\includegraphics[width=6.2cm]{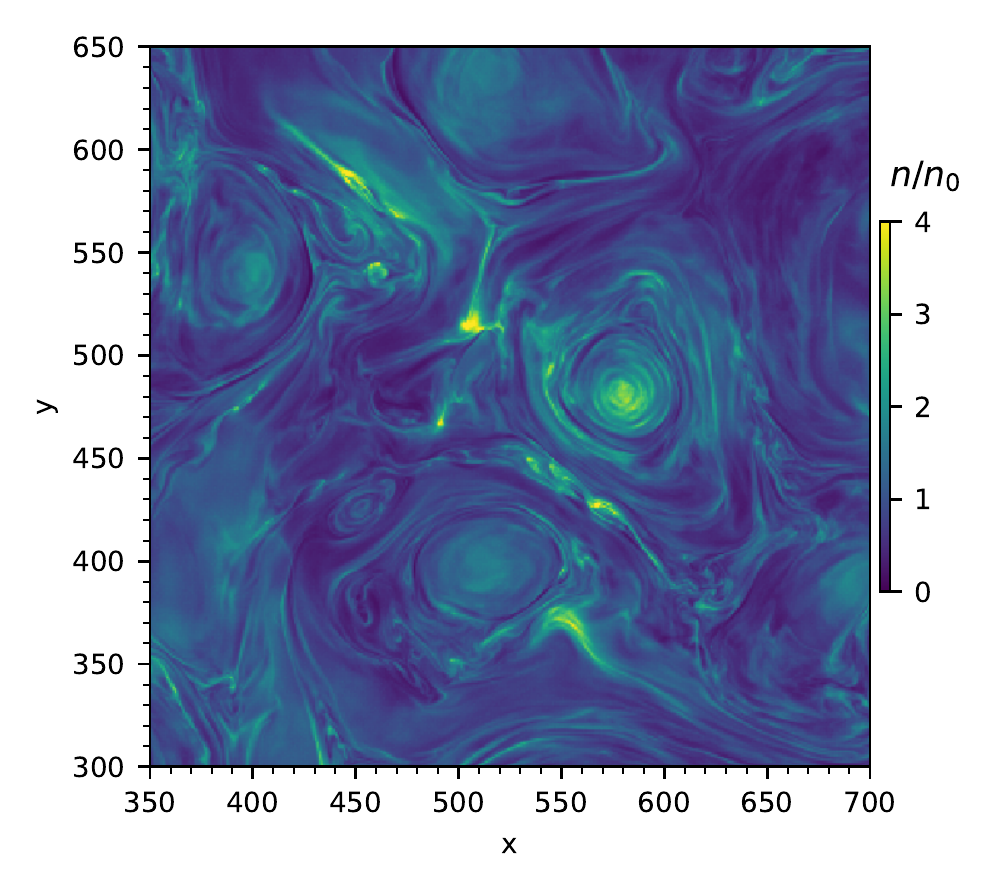}
    \centering\includegraphics[width=6.2cm]{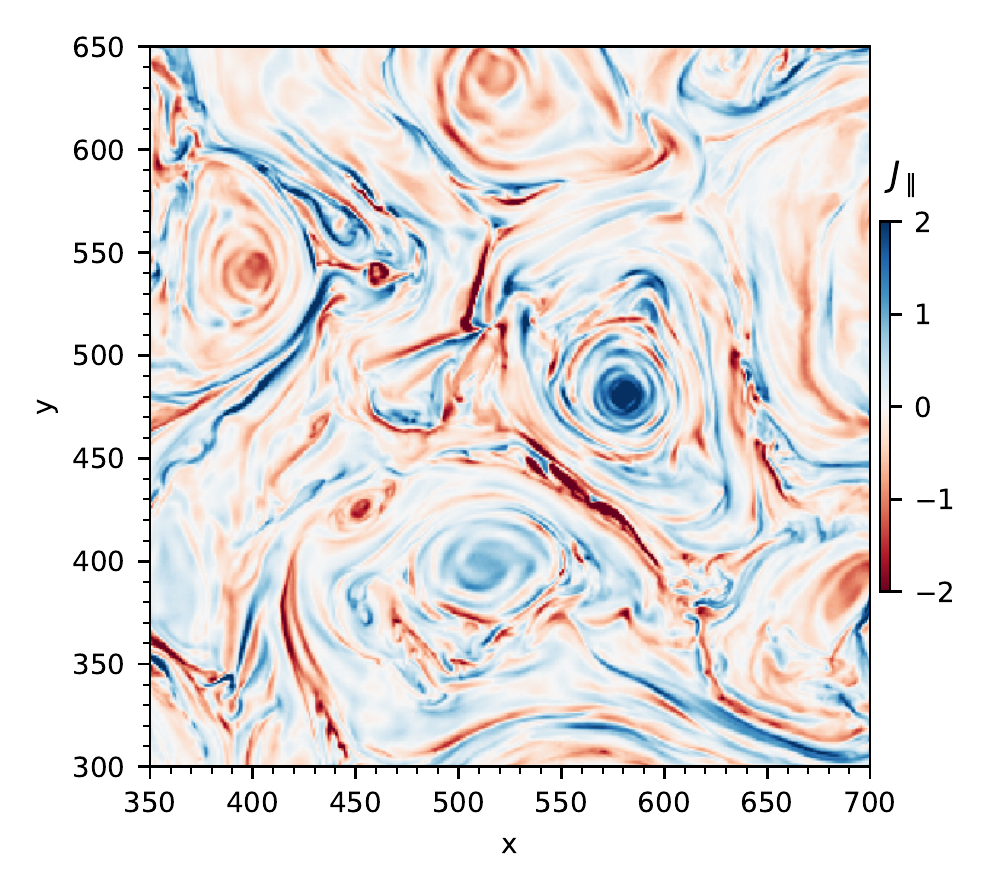}
    \centering\includegraphics[width=6.2cm]{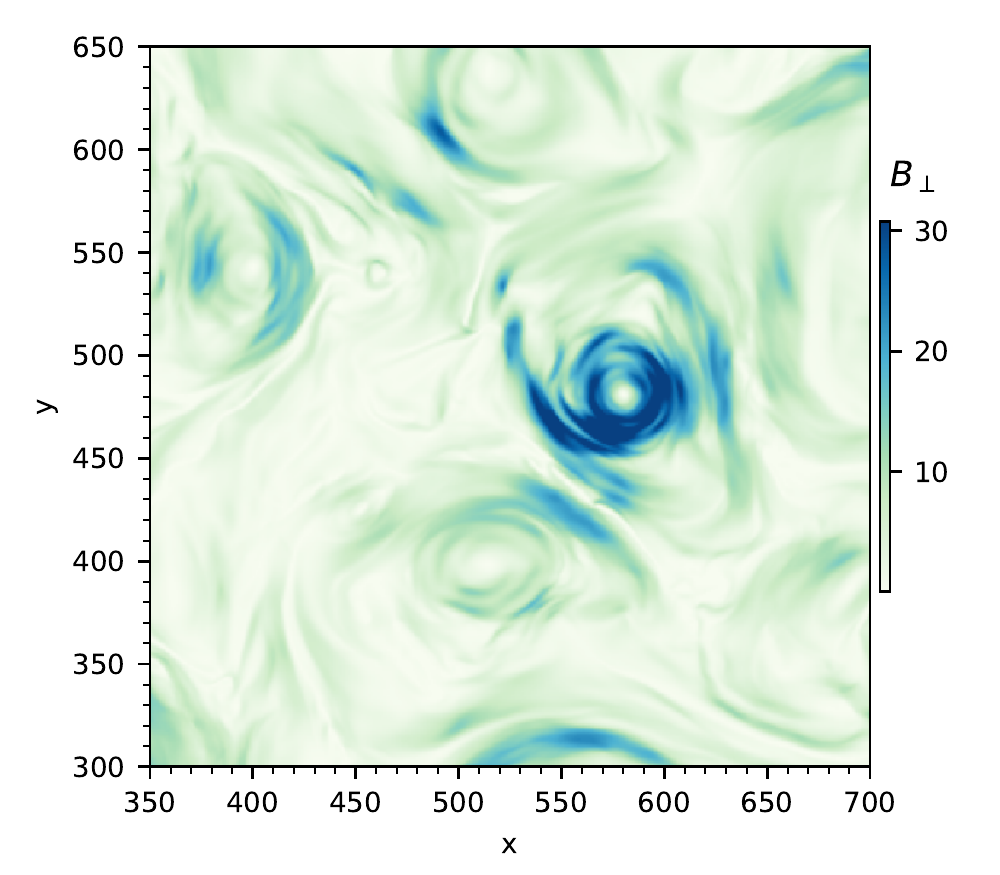}
    \centering\includegraphics[width=6.2cm]{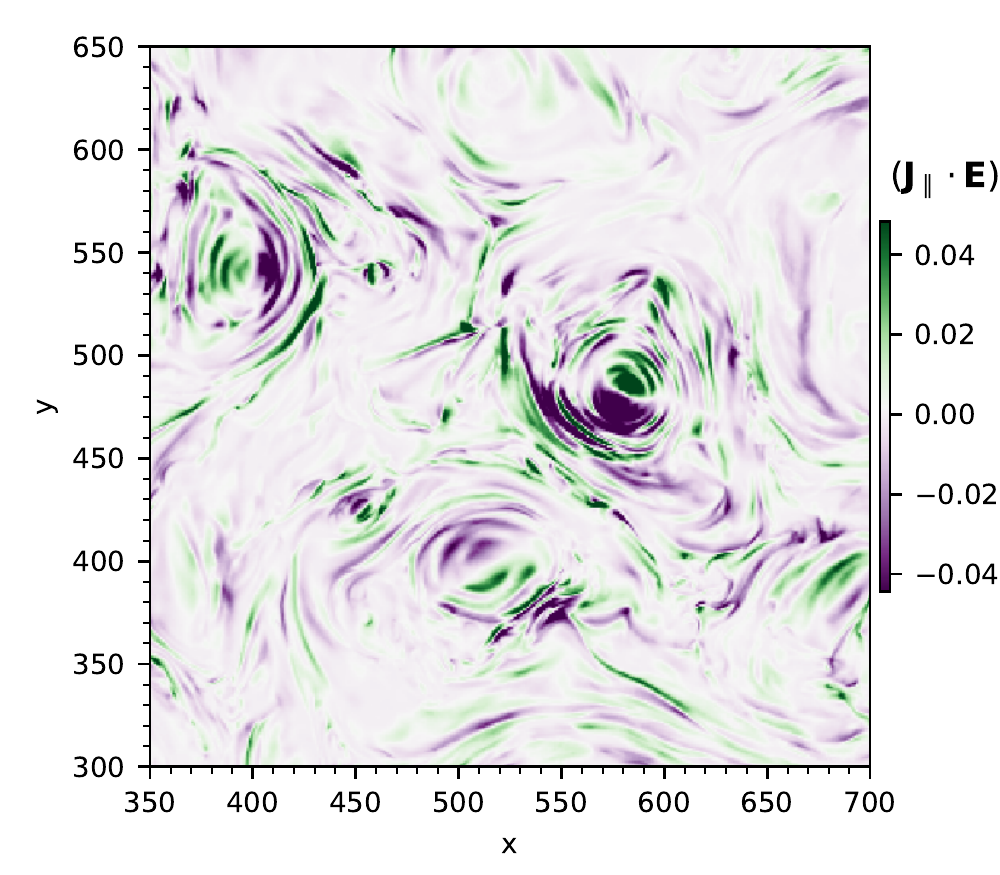}
      \centering\includegraphics[width=6.2cm]{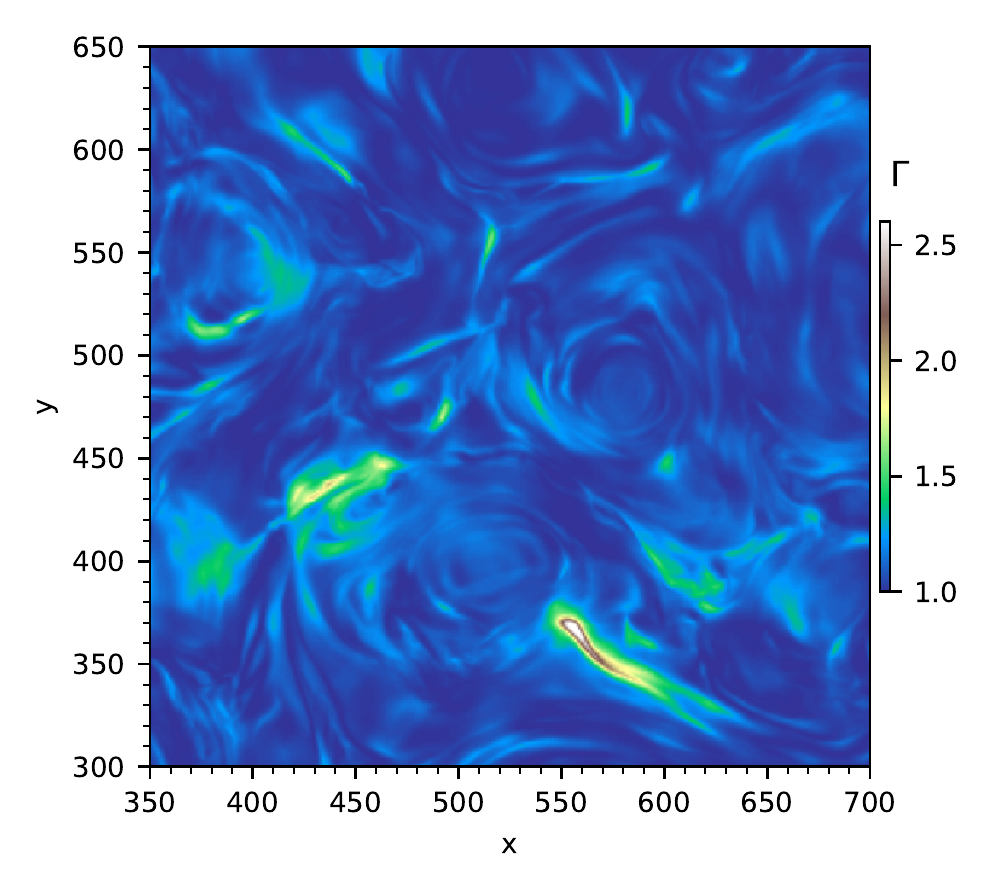}
  \caption{
    Visualization of the turbulent plasma simulation data. 
    Each pixel has four different features that we visualize here:
    $n/n_0$
    (top left), 
    zoom-in 
    of $n/n_0$
    (top right panel),
    $J_{\parallel}$ (middle left panel),
    $B_{\perp}$ (middle right panel),
    $(J_\parallel \cdot E)$ (bottom left panel) and
    $\Gamma$
    (bottom right panel.)
    The top left panel shows the full image domain whereas the rest of panels show a close-up region of the image (marked with red-dashed line in the top left panel).
}
\label{IMG:PhysicalStructures}
\end{figure*}

Figure \ref{IMG:PhysicalStructures} visualizes the used simulation data.  
The visualizations shows five physical features processed from the raw simulation data. 
The upper left panel shows the whole simulation box  consisting of a little over $1$ million pixels. 
The depicted feature is the plasma number density $n$ (in units of initial number density $n_0$) at a physical time of $t \approx 5$ eddy turnover times.
Rest of the images are zoom-in views to the simulation box. 
Left panel of the second row shows the current along the out-of-the-plane $z$-axis, $J_{\parallel}$. 
Right panel of the second row shows the magnetic field strength perpendicular to the out-of-plane $z$-axis, $B_{\perp}$.
The lower left panel visualize the work done by the parallel electric field $(J_\parallel \cdot E)$ and lower right panel the plasma bulk Lorenz factor, $\Gamma$.
All of these visualized features have multiple self-similar structures that are clearly visible in Fig. \ref{IMG:PhysicalStructures}.
For example, we identify circular islands that coincide with a high signal in $n/n_0$, $B_{\perp}$ and in $J_{\parallel}$. 
Another prominent feature are the currents sheets that correspond to a maximum in $J_{\parallel}$ and $\Gamma$, and a minimum in $B_{\perp}$.

\section{SOM algorithm}\label{sec:SOMtheory} 

We start by presenting a short overview of the SOM algorithm and its process parameters, following  \citet{kohonen2013essentials}, \citet{Kohonen1989} and \citet{kohonenSOM}.
Next, we discuss some of its shortcomings that motivated the development of the subsequent stacking framework.

\subsection{Theory}

The data sample is described by a vector $X=\{X_{1}$, $\ldots$, $X_{k}\}$, where each element $X_{i}$ describes a set of input variables $\xi_j$, such that $X_{i}=[\xi_{1}$, $\dots$, $\xi_{l}] \in \mathcal{R}^{l}$, $\forall i \in \{1,\dots, k\}$. 
We adopt a regular 2-dimensional rectangular shaped Kohonen neural map of dimensions $(m,n)$. 
\citet{kohonen2013essentials} advises to select map dimensions such that they describe the lengths of the first principal components and for a bigger map size to detect fine structures.
A rigorous choice for the dimensions and architecture of the map will result in a faster convergence.

Each neuron on the map is associated with a parametric vector $w_{i}=[\mu_{1},\dots, \mu_{l}] \in \mathcal{R}^{l}$. 
The initial values for the parametric vectors of nodes may be sampled randomly from the domain of the input vector parameter space.
A more sophisticated nomination could also be used, if needed \citep{kohonen2013essentials,valova2013initialization}.

Let $d(X_{s}, w_{i})$ denote the distance metric used to determine the Best Matching Unit, $w_{BMU}$ (BMU) from the 2-dimensional neuron map, which is the neuron most similar to the sampled data vector $X_{s}$,
\begin{equation}
    w_{BMU}=argmin_{i}\{d(X_{s},w_{i})\} \label{BMU}.
\end{equation}
In this work we use the Euclidean distance on normalized data vectors $X_s$ and neuron models $w_i$, but other measures can also be used. 
The data vector $X_{s}$ is compared to all the neurons $w_{i}$. 
The neuron with the smallest Euclidean distance is chosen as the BMU.

A neighborhood $N_{c}$ consists of all the nodes up to a certain distance $r$ from a node $w_{c}$ on the neuron map, according to some geometric distance. 
The set usually shrinks with time and is determined by the neighborhood function $h_{ci}(t)$.
The choice of $N_{c}$ influences the map's ability to order and learn the underlying data distribution 
\citep{kohonenSOM, lee2002self}. 
In the SOM algorithm learning step, both the BMU and its spatial neighbors $N_{BMU}$ learn from the input vector.  

The learning step to modify the neuron weight vectors is given as\\
\begin{equation}\label{learning}
    w_{i}(t+1) =
  \begin{cases}
     w_{i}(t) + \alpha(t) * h_{ci}(t)[X_{s}(t) - w_{i}(t)],\quad w_{i} \in N_{c}, \\
     w_{i}(t), \quad w_{i} \notin N_{c},
    \end{cases}
\end{equation}
where $t=0,1,2,\dots$ is a discrete time value. 
The value $0 < \alpha(t) < 1$ is the learning rate, which determines the statistical accuracy of the neuron map and the ordering of the neurons on the map. 
The function $h_{ci}(t)$ acts like a neighborhood function and for convergence $h_{ci}(t) \rightarrow 0$ as $t \rightarrow \infty$.
Neighborhood function determines the rate of change for neurons on the map \citep{kohonenSOM, kohonen2013essentials, stefanovivc2011influence}.

Neurons in the neighborhood $N_{BMU}$ of the winning neuron $w_{BMU}$ will be updated in accordance to a chosen criterion during the learning step. 
During this the smoothing of the map takes place. 
The matching law used in Equation~(\ref{BMU}) and the updating law in Equation~(\ref{learning}) need to be compatible.
 
The algorithm is stochastic, which means that the reliability of the learning is also dependent on the number of training steps. 
\citet{kohonenSOM} proposes an empirical rule of thumb of $500$ times the count of network units (neuron map size) for the total number of training steps.

The result of the SOM is a $2-$dimensional neural map representing the $l-$dimensional input space.
Each input sample vector $X_{s}$ has a neuron on the neuron map, which describes its parametric vector the best on the two-dim\-en\-sio\-nal map. 

\subsection{Stability of SOM clustering}\label{SEC:SOMConfidence}

The obtained clusters depend on the initial neuron map size ($m\times n$), learning rate $\alpha(t)$, neighborhood function $h_{ci}(t)$, training step size, distance metric, the rule for joining the neurons, sampling of the input vector $X_{s}$, the nature of initialization of the neurons and the stochastic nature of the algorithm. 
Thus two SOM algorithm evaluations on the same data, applying same process parameters and initial functions, commonly converge to somewhat different clustering results \citep{statisticsSOM}. In addition we do not actually know, whether the algorithm has reached a global optimum.

We apply the SOM clustering method to astrophysical plasma simulation data 'images'. 
Our goal is to use a clustering algorithm to detect different physical structures in the data.
We observe that the results obtained with the SOM algorithm are strongly dependent on the selection of process parameters, iteration steps and randomness of the initial conditions. 
This volatility of the segmentation  is visualized in Figure~\ref{IMG:SOMResults} showing four independent SOM runs with only slightly different process parameter values.
Notably, we obtain differing cluster sets for the same input image.

\section{Stacking of SOMs in the SCE framework}\label{atlastheory}

We will now present a new theoretical description for stacking of many clustering evaluations in order to obtain more stable clustering regions.
We call the algorithm the Statistically Combined Ensemble (SCE) since it is based on clustering results in not one map but in a statistically combined ensemble of independent maps.
We emphasize that the method is not limited to the use of SOMs as the base algorithm. 
The SCE ensemble framework is specifically designed for unsupervised image segmentation ensembles. 

In this case study, first we obtain $N$ independent SOM clustering results for astrophysical simulation image data. 
The SCE framework stacks cluster maps detected in independent SOM algorithm runs and gives an estimate to how well the observed cluster fits with other clusters detecting a similar structure.
A key realization that enables to combine different SOM results is that we use the projected original images to perform the stacking operations;
not the SOM results themselves that are bound to change from one realization to another.
Metrics to estimate the goodness of fit between cluster maps are then constructed. 
As a result we obtain cluster elements that are the best detected in all the independent SOM runs.

\subsection{Cluster masks}

Applying the SOM algorithm to an image of size $r\times t$ will result in dividing the original data into $n$ clusters.
Each pixel $p_{ij}$, where $i \in \{1,\dots, r\}$ and $j \in \{1,\dots, t\}$, on this image can be associated with a cluster from the set of detected $n$ clusters $\{C_1,\dots, C_{n}\}$, where $n, C_1, C_{n} \in \mathbb{N}$.  

For every cluster $C_{k}$ in the set of clusters $\{C_{1}, \dots, C_{n}\}$ a mask $M_{k}$ can be defined.
A mask is a boolean matrix $M_{k}=(m_{i,j}): r \times t$, such that
\begin{equation}
    m_{ij} =
  \begin{cases}
    0         & \quad \text{if }  p_{ij} \notin \text{ cluster } C_{k},\\
    1         & \quad \text{if }  p_{ij} \in \text{ cluster } C_{k}. \label{mask}
    \end{cases}
\end{equation}
Elements in this mask matrix $M_{k}$ consists of values $1$ if the pixel is in the observed cluster $C_{k}$ and $0$ otherwise. 
Using the set of clusters $\{C_1, \dots, C_{n}\}$ we have now defined $n$ mask matrices $M_{1},\dots, M_{n} : r \times t$.
The mask matrix describes the locations and area of the structure on the image providing a mapping from the cluster groupings into the initial data view.

We can then apply the SOM algorithm $N$ independent times on the same image data. 
This means that for each pixel $p_{ij}$ on the image we will have $N$ independent clusters. 
In other words, we have gained $N$ cluster sets $\{C_1,\dots,C_{n_1}\}, \dots, \{C_1,\dots, C_{n_N}\}$, with $n_{1}, \dots n_{N}$ elements.
From these, we can then create $N$ independent sets of mask matrices $\mathcal{M}=\Big\{\{M_{1}^{1}$, $\dots$, $M_{n_{1}}^{1}\}$, $ \dots$, $ \{M_{1}^{N}$, $\dots$, $ M_{n_{N}}^{N}\}\Big\}$ according to Equation~(\ref{mask}). 
We use a notation where the upper index of a mask matrix refers to the SOM realization index and the lower index to the detected cluster in that realization. 

\subsection{Mask-to-mask stacking}\label{masktomask}

Let us randomly select a set of masks $\mathcal{M}^{b}=\{M_{1}^{b}$, $ \dots$, $ M_{n{_b}}^{b}\}$, $1\leq b \leq N$, from the set of independent mask sets $\mathcal{M}$. 
We call this mask set the base mask set. 
We compare each mask in $\mathcal{M}^{b}$ to every other mask in the set $\mathcal{M}\setminus \mathcal{M}^{b}$. 
For every mask in $\mathcal{M}^{b}$ we obtain $n_{1}+n_{2}+\dots + n_{N} - n_{b}$ comparisons. 
The comparison is performed between every mask in a randomly chosen base mask set against every other mask set independent of $\mathcal{M}^{b}$. 
This process is carried out as long as each set of masks has been chosen as the base mask set.  

In supervised segmentation the obtained result is compared to the known truth about the data. 
In the 
unsupervised case the base mask matrix is compared to every other cluster realization from $\mathcal{M}\setminus \mathcal{M}^{b}$. 
The stacking of masks from $\mathcal{M}\setminus \mathcal{M}^{b}$ on the base mask from $\mathcal{M}^{b}$ is used to evaluate the reliability of the base mask.

\subsubsection{Similarity measures}

We define three different similarity measure matrices for a mask $M_{e}^{b}$ in the base mask set $\mathcal{M}^{b}$, with $n_{b}$ masks and  $1 \leq e \leq n_{b}$, and a mask $M_{f}^{c}$ from a mask set $\mathcal{M}^{c}$, with $n_{c}$ masks, $1 \leq f \leq n_{c}$ and $b \neq c$.
These matrices are:
The union matrix $U: r \times t$, with possible values of $\{0,1\}$, defined as
\begin{equation}\label{EQ:union}
    U(i,j) =
  \begin{cases}
    0        & \quad \text{if }  M_{e}^{b}(i,j)=0 \wedge  M_{f}^{c}(i,j)=0 \\
    1     & \quad \text{otherwise}, 
  \end{cases}
\end{equation}
where $i \in \{1,\dots, r\}$ and $j \in \{1,\dots, t\}$.
The intersection matrix $I: r \times t$, with possible values of $\{0,1\}$, defined as
\begin{equation}\label{EQ:intersection}
    I(i,j) =
  \begin{cases}
    1         & \quad \text{if }  M_{e}^{b}(i,j)=1 \wedge  M_{f}^{c}(i,j)=1 \\
    0        & \quad \text{otherwise},  
  \end{cases}
\end{equation}
where $i \in \{1,\dots, r\}$ and $j \in \{1,\dots, t\}$.
The sum matrix $R: r \times t$, with possible values of $\{0,1,2\}$, is defined as
\begin{equation}\label{EQ:sum}
    R(i,j) =
  \begin{cases}
    0        & \quad \text{if }  M_{e}^{b}(i,j)=0 \wedge  M_{f}^{c}(i,j)=0 \\
    1      & \quad \text{if } M_{e}^{b}(i,j)=1 \wedge M_{f}^{c}(i,j)=0 \\
              & \quad \vee M_{e}^{b}(i,j)=0 \wedge M_{f}^{c}(i,j)=1 \\ 
    2       & \quad  M_{e}^{b}(i,j)=1 \wedge M_{f}^{c}(i,j)=1, 
  \end{cases}
\end{equation}
where $i \in \{1,\dots, r\}$ and $j \in \{1,\dots, t\}$.\\
We calculate these similarity measures between mask $M_{e}^{b}$ in the base mask set $\mathcal{M}^{b}$ and every mask in the set $ \mathcal{M} \setminus \mathcal{M}^{b}$.
This is done until all masks in the set $\mathcal{M}$ have been chosen as a base mask set.

\subsubsection{Signal strength}\label{Sect:signalstrength}

Using the similarity matrices we can construct 
a so-called signal strength $s_I$ measure.
It quantifies how well the observed cluster resembles other clusters.
We denote two independent clusters similar, if their mask matrices are identical.

The signal strength metric $s_I$ is identical to the Mean Intersection-over-Union (MIoU) metric commonly used in supervised segmentation. 
In the supervised learning the MIoU metric is used to compare the detected segmented objects to the ground truth ---
information that is lacking in the unsupervised learning case.  
The $s_I$ metric is defined between independent realizations of unsupervised clustering algorithms. 
Every evaluation set is taken as the ground truth and each of these "true" cluster masks matrices are compared to every other cluster result from independent realizations. 

The signal strength gives an estimate for the strength of intersection of the mask matrices, measuring the intersection in comparison to the union of the two masks.
For any base mask matrix $M_{e}^{b}$ and a random mask $M_{f}^{c}$, where $b \neq c$, the signal strength scalar $s_{I}$ is defined as
\begin{equation}
    s_{I} = \frac{\sum_{i,j=1}^{r\times t} I(i,j)} {\sum_{i,j=1}^{r\times t} U(i,j)}. \label{EQ:signalscalar}
\end{equation}

The signal strength scalar $s_I$ weighs the intersection matrix (Equation~(\ref{EQ:intersection})) of two masks with their union matrix (Equation~(\ref{EQ:union})). 
This is done to eliminate the cluster size dependency; 
large clusters will likely have more common pixels with any other clusters in independent SOM runs, so we need to counter this effect by normalizing the quantities with the union measure.

If $\sum_{i,j=1}^{r\times t} I(i,j) \rightarrow \sum_{i,j=1}^{r\times t} U(i,j)$, then $s_{I} \rightarrow 1$. 
This means that the two masks have the value $1$ in the same locations and in same quantity in their $r \times t$ matrices. 
In that case, these two independent SOM algorithm evaluations have detected the same shape structures in the same pixel locations. 

The signal strength scalar $s_{I}$ is a normalized quantity, which makes it a quantitative measure to order and rank different masks. 
The mask $M_{e}^{b}$ will have $n_{1}+n_{2}+\dots + n_{N}-n_{b}$ elements in the vector of all signal strength scalars, a vector noted as $\overrightarrow{s_{I}}$. 
Each mask in the set of all masks $\mathcal{M}$ will obtain a signal strength vector. 
Using this measure we can order the masks.

\subsubsection{Quality of cluster unions}\label{Sect:quality}

Another accompanying measure to the strength is the so-called quality measure as it gives an estimate to the quality of the union of the stacked masks. 
The best quality measure for a specific cluster corresponds to having a union of two masks close to their intersection.
This corresponds to a small residual area, area of the symmetric difference in set theoretical notion, between the masks.
This means that their mask matrices overlap perfectly resulting in a union that is equal to the intersection. 
The worst quality measure, on the other hand, corresponds to a case when the area of the union of two masks is close to the area of their sum. 

For any base mask matrix $M_{e}^{b}$ and a random mask $M_{f}^{c}$, where $b \neq c$, the quality scalar $q_{U}$ is defined as
\begin{equation}
    q_{U} = \frac{\sum_{i,j=1}^{r\times t}U(i,j)} {\sum_{i,j=1}^{r\times t}R(i,j)}-
    \frac{\sum_{i,j=1}^{r\times t} I(i,j)} {\sum_{i,j=1}^{r\times t} R(i,j)}. \label{EQ:qualityscalar}
\end{equation}
The value of quality scalar $q_{U}$ goes to zero as $\sum_{i,j=1}^{r\times t} U(i,j)$ $\rightarrow \sum_{i,j=1}^{r\times t} I(i,j)$, which means the two independent stacked masks $M_{e}^{b}$ and $M_{f}^{c}$ have detected exactly the same structure in the same locations on the image. 
On the other hand, as $q_{U}$ goes to unity, the element sum of the union matrix $U(i,j)$ and element sum of the sum matrix $R(i,j)$ is equal. 
This means that the intersection of the two masks is negligible and they correspond to completely different structures. 
The union quality measure $q_{U}$ can be used to discard clusters from independent SOM runs that have detected completely different structures from other evaluations. 

This metric is similar, but not equivalent, to the Dice coefficient that is widely used in the supervised image segmentation.
Using the intersection and union matrices defined in this paper, the Dice coefficient is defined as $D=\frac{2 \cdot \sum_{i,j=1}^{r\times t} I(i,j)}{\sum_{i,j=1}^{r\times t} R(i,j)}$.
The Dice coefficient is positively correlated with the MIoU whereas the $q_U$ of cluster masks is negatively correlated to the MIoU, or $s_I$ in our concept.

\subsection{Stacking of multiple masks}\label{SEC:multiplemasks}

In this subsection we will generalize the mask-to-mask comparisons to general quantities averaged over the complete set of independent realizations.
To do this, we combine the signal strength  $s_{I}$ and quality of union $q_{U}$. 
We will denote this matrix with $G$, as it estimates the goodness of the fit between independent cluster masks. 
The measure peaks in areas, where the base mask has often a high percentage of its area in common with other detection algorithm clusters. 
This results in highlighting the highest probability structures on the image. 
The measure gives a quantitative value to estimate the goodness of fit between independently detected cluster masks. 
Cluster masks, which have detected the same structures, will fit together well and contribute more to the total sum of the measure.
Similarly, cluster masks that do not represent the same physical structures end up not contributing significantly to the total integrated goodness measure for that base mask. 

In subsections~\ref{Sect:signalstrength} and~\ref{Sect:quality} a mask $M_{e}^{b}$ in the base mask set $\mathcal{M}^{b}$ is combined with every mask in the set $ \mathcal{M} \setminus \mathcal{M}^{b}$. 
This corresponds to $n_{1}+n_{2}+\dots + n_{N}-n_{b}$ signal strengths $s_{I}$ and qualities of union $q_{U}$. 
The goodness of fit of a cluster mask $M_{e}^{b}$ to any other cluster mask $M_{f}^{c}$ from $ \mathcal{M} \setminus \mathcal{M}^{b}$, is defined as
\begin{equation}
    G=
    \frac{s_{I}}{q_{U}} \cdot
    (M_{e}^{b} \cup M_{f}^{c}) \label{EQ:SQ},
\end{equation}
where $s_{I}$ and $q_{U}$ are the signal strength scalar and quality of union scalar of the masks $M_{e}^{b}$ and $M_{f}^{c}$.
The matrix $M_{e}^{b} \cup M_{f}^{c}$ refers to all the  pixels on the image that belong to either of these two cluster masks.
For an observed mask in the base mask set $\mathcal{M}^{b}$ we obtain $n_{1}+n_{2}+\dots + n_{N}-n_{b}$ matrices according to the Equation~(\ref{EQ:SQ}), which have the corresponding quotient value in the union of the two compared masks. 
Then for every base mask $M_{e}^{b}$ in $\mathcal{M}^{b}$ all $n_{1}+n_{2}+\dots + n_{N}-n_{b}$ of its $G$ matrices can be summed together to yield
\begin{equation}
    G_{\mathrm{sum}}=
   \sum_{k=1}^{N}\sum_{l=1}^{n_{k}} \Big(\frac{s_{I}}{q_{U}} \cdot
    (M_{e}^{b} \cup M_{l}^{k}) \Big) \label{EQ:SQ_sum},
\end{equation}
where $k \neq b$. 
This is done for every base mask, until all mask sets in $\mathcal{M}$ have been chosen as a base mask set.
We then obtain $n_{1}+n_{2}+\dots + n_{N}$ summed goodness of fit matrices $G_{\mathrm{sum}}$. 
Each of them characterizing the fit of a cluster mask to all other cluster masks detected in other independent SOM cluster evaluations. 

By summing the quotients of the signal strength and the union quality in the Equation~(\ref{EQ:SQ_sum}), we will accumulate value to pixels, where the base mask and other independent masks have detected a structure. 
The value added to the $G_{\mathrm{sum}}$ by a base mask and a compared mask will be high for those pairs, that have a similar number of pixels assigned to the cluster and their masks have value $1$ in same location. 
The contribution from mask pairs, which detect distinct structures on the image, is negligible since the $\frac{s_{I}}{q_{U}}$ will be close to zero in value for those cases. 

We apply the stacking framework on clusters detected by the SOM algorithm. 
From SOM results we attain in total $n_{1}+n_{2}+\dots n_{N}$ cluster masks. 
After calculating the $G_{\mathrm{sum}}$ for a randomly chosen base mask $\mathcal{M}_{e}^{b}$, each pixel on the image obtains a value between $0$ and $1$ illustrating its stability of belonging to the cluster with that shape and location. 
In a sense every cluster realization from all of the SOM realizations is viewed as the ground truth for a cluster.

Next we define a scalar value for the goodness of fit for $\mathcal{M}_{e}^{b}$ as the following
\begin{equation}\label{EQ:gsumscalar}
g_{\mathrm{sum}}=\sum_{\mathcal{M}\setminus \mathcal{M}^{b}} (\frac{s_{I}}{q_{U}}),
\end{equation}
where the sum is taken over all the comparisons between $\mathcal{M}_{e}^{b}$ and every other mask in $\mathcal{M}\setminus \mathcal{M}^{b}$. 

By ordering the scalar $g_{\mathrm{sum}}$ of all cluster masks, we can find the base mask, which has learned most of the information that describes this cluster. 
In other words, this mask fits the best with all the other independent realizations of that cluster. 
Same cluster realization base masks will likely follow the winning base mask stack on the ordering of the $g_{\mathrm{sum}}$ value.
Correlated classifiers for clusters from an unsupervised ensemble framework indicate an accurate classification \cite{pmlr-v51-jaffe16, Platanios2017, ROKACH2009}. 
A base mask detecting a cluster from the image is a classifier for a given object.
This means that cluster masks with $g_{\mathrm{sum}}$ values that are highly correlated indicate that the cluster realization is accurate.

Therefore, same clusters are bound to have similar goodness of fit measures;
a large change in its value indicates a physically different cluster group or an non-accurate classifier for the given group.

\subsection{Summary of the SCE algorithm}

\begin{figure*}
   \centering
   \includegraphics[width=16.2cm]{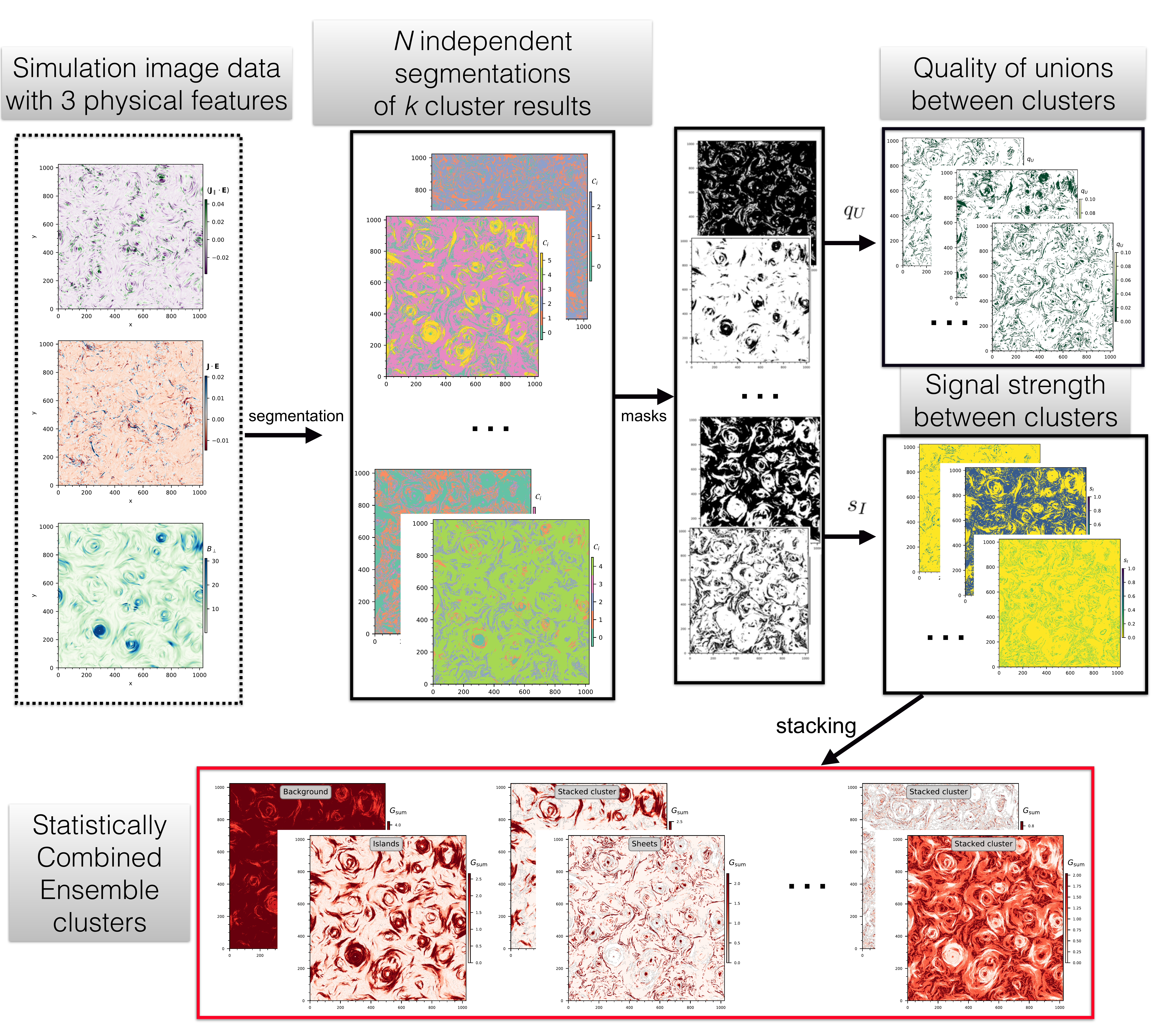}
     \caption{
    Step-wise description of the SCE framework. 
    The framework 
    (\textit{i}) takes images as an input, 
    (\textit{ii}) performs multiple independent segmentations of them, 
    (\textit{iii}) combines the different segmentation results into so-called quality and signal strength measures, and
    (\textit{iv}) outputs statistically combined ensemble averages of these realizations. }
     \label{IMG: Systematic plot}
\end{figure*}

SCE framework is summarized in Figure~\ref{IMG: Systematic plot}.
The input data is a set of simulation images obtained from an astrophysical simulation, each pixel has 3 different physical features attached. These images are fed into N independent unsupervised clustering algorithms, to obtain $\{C_1,\dots,C_{n_1}\}, \dots, \{C_1,\dots, C_{n_N}\}$ cluster realizations, with $n_{1}, \dots n_{N}$ elements (in this work the Self-Organizing Map was used). Clustering results are then deconstructed into sets of mask matrices
\begin{equation}
    \mathcal{M}=\Big\{\{M_{1}^{1},\dots,M_{n_{1}}^{1}\}, \dots, \{M_{1}^{N},\dots, M_{n_{N}}^{N}\}\Big\},\nonumber 
\end{equation}
where each mask $\mathrm{M}$ represent a cluster detected in the image in the specific clustering run. These masks are then compared with all other independent cluster masks with a quality of union metric and signal strength metric. These similarity measures are combined together to create a goodness-of-fit metric the $G$. Then for each cluster mask all  $n_{1}+n_{2}+\dots + n_{N}-n_{b}$ of its $G$ matrices are summed together to yield the $G_{sum}$ matrix. These stacked matrices for each detected cluster are the output of the SCE framework. 

\section{Applying SOM and SCE}\label{SEC:Application}

In this section we apply the SOM and then the SCE method on complex images of turbulent magnetically dominated collisionless plasma simulations described in Section \ref{SEC:data}.
The images we dissect ---
and that inspired us to develop the discussed method --- 
are 2-dimensional computer simulation images.
Our main aim is to dissect the simulation snapshot pixels into distinct physical structures based on the similarity of their feature vectors.
We are, in particular, interested in automating the grouping of pixels into three different physical shapes: magnetic flux tubes (islands), current sheets (long stripes), and background medium \citep[see e.g.,][for a more detailed discussion of the physics]{Zhdankin2017, Comisso2018, Nattila2019}.

Similar problem setups can also be envisioned in many other fields of science like, for example, in medicine where medical images need to be dissected and segmented into different structures automatically.

\subsection{Results of SOM segmentation}

We apply the SOM algorithm to $8$ consecutive simulation image snapshots.
We use rectangular shaped neuron map with dimensions $(15,10)$.
The learning rate values are chosen to be $0.6$, $0.7$, or $0.8$.   
Number of total iterations is chosen to be $10~000$, $20~000$, $30~000$, $40~000$, or $50~000$.

\begin{figure*}
   \centering
    \centering\includegraphics[width=6.2cm]{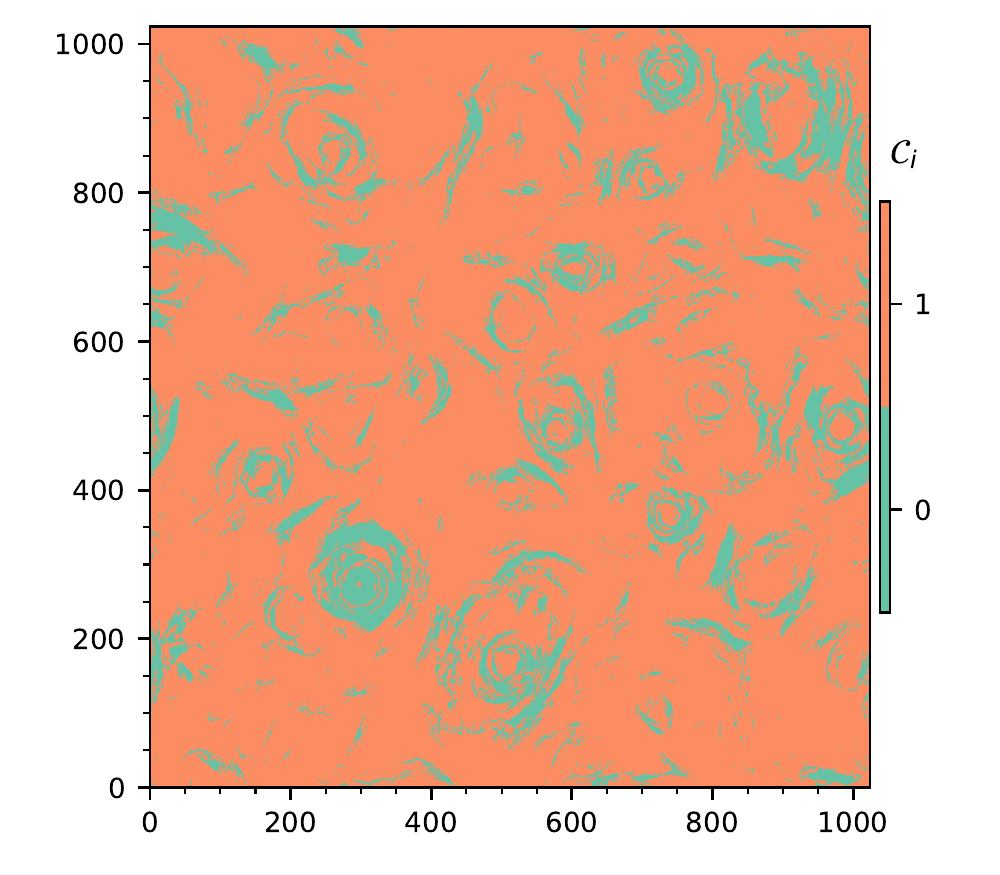}
    \centering\includegraphics[width=6.2cm]{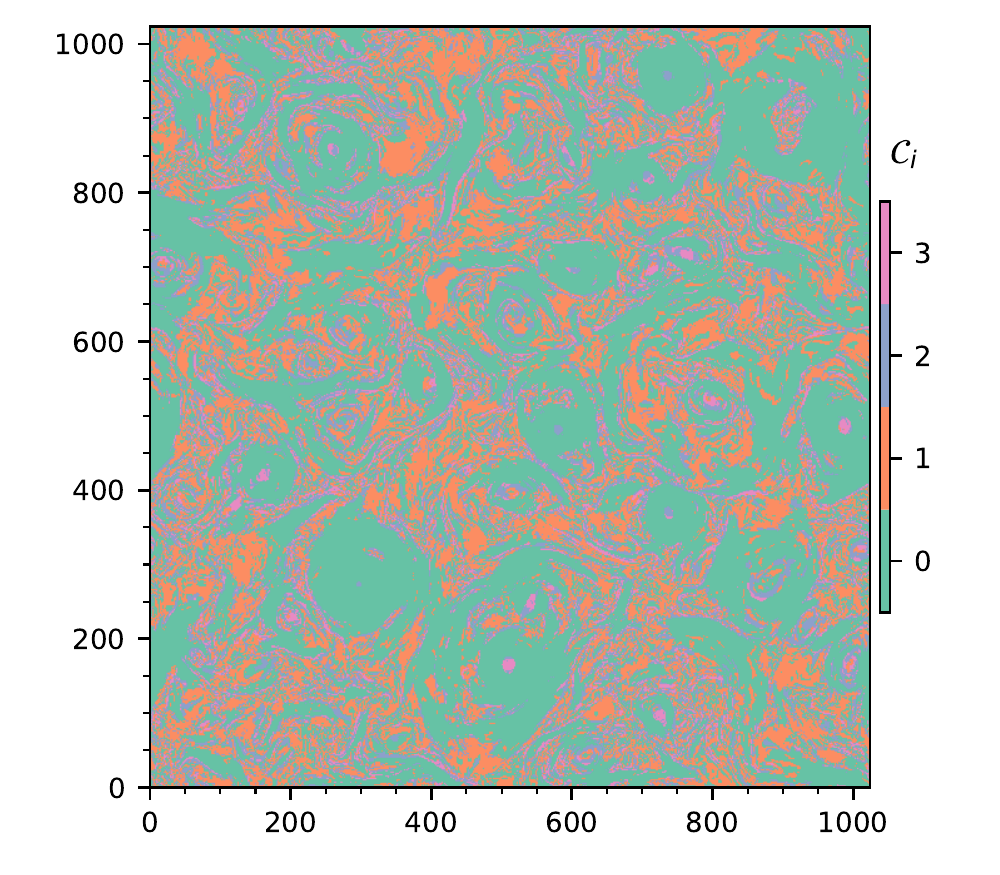}
    \centering\includegraphics[width=6.2cm]{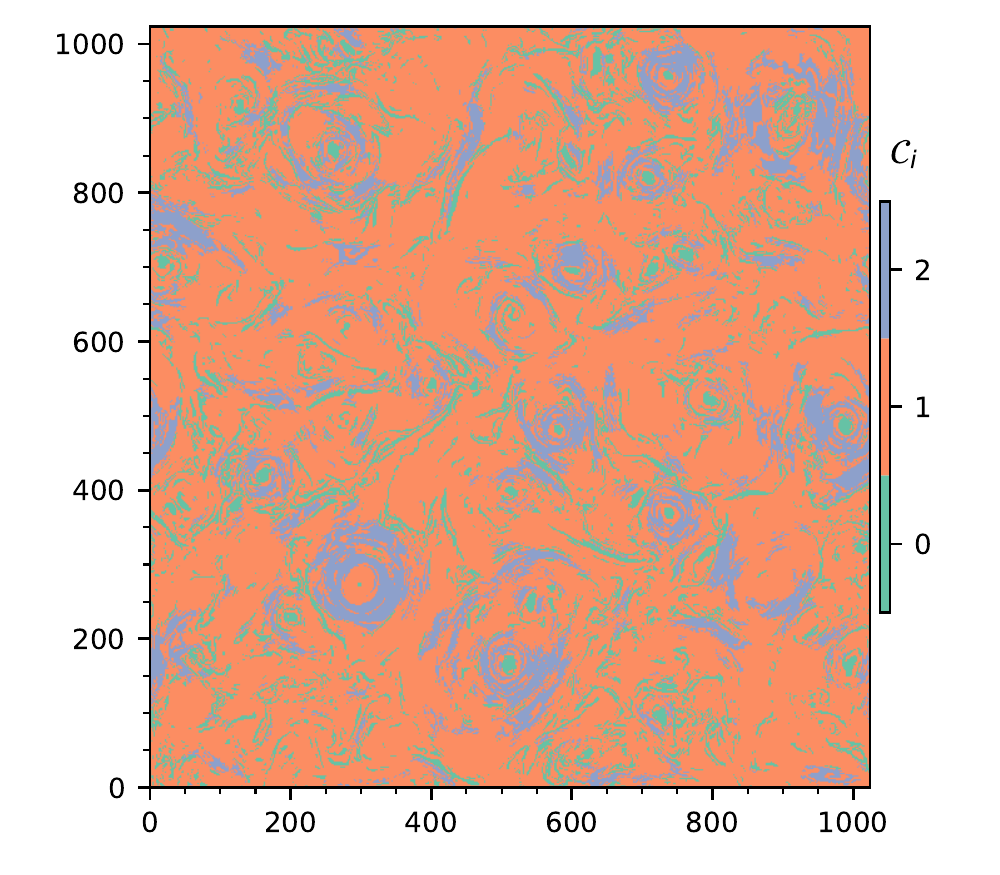}    
    \centering\includegraphics[width=6.2cm]{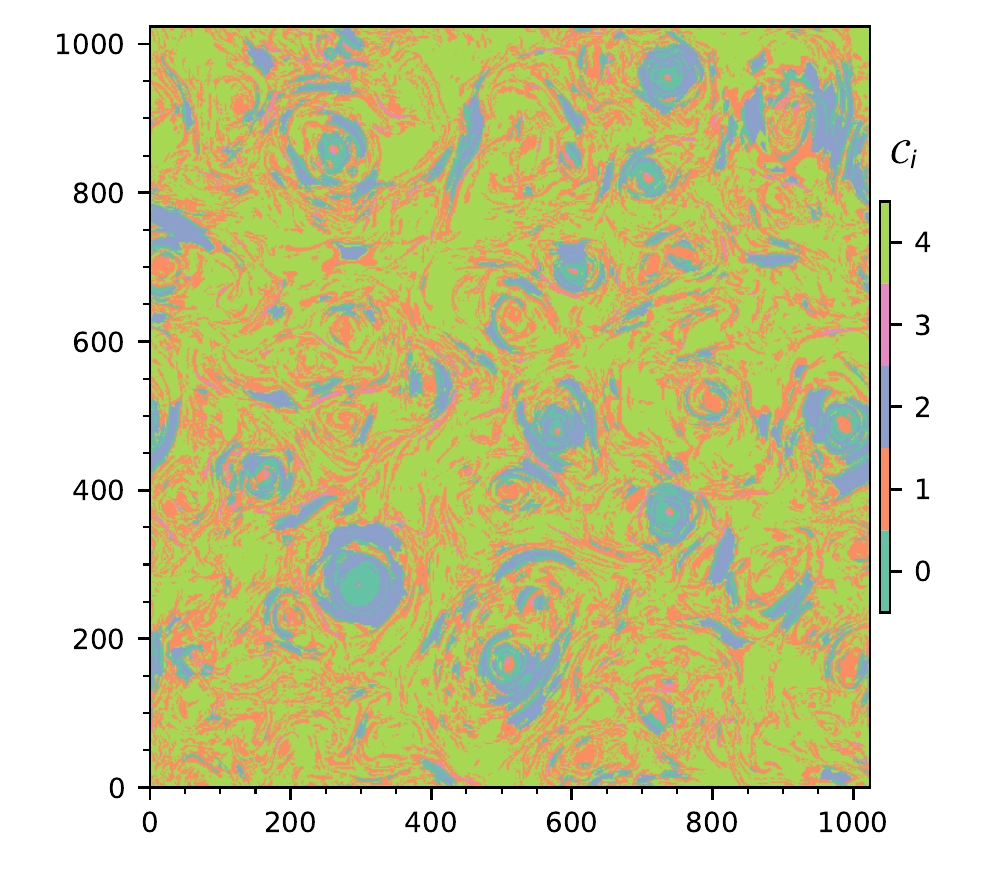}
  \caption{
  Visualization of four different SOM evaluations projected back to the original image view.
  Each of the SOM runs group the initial data pixels into clusters, $\mathcal{C}_i$.
  The pixels are colored based on the cluster it is associated with in order to highlight the physical structures that the data clusters correspond to. 
  }\label{IMG:SOMResults}
\end{figure*}

SOM algorithm is applied on the studied images with all the possible aforementioned parameter combinations. 
This gives $15$ independent SOM clustering results for the astrophysical plasma simulation images. 
Figure~\ref{IMG:SOMResults} visualizes results from four representative SOM clustering algorithm realizations projected back to the original image view. 
The resulting clusters can be directly compared to Figure~\ref{IMG:PhysicalStructures}.
The clusters detected by the SOM algorithm on Figure~\ref{IMG:SOMResults} correspond to distinct regions in the original images that are also visible in Figure~\ref{IMG:PhysicalStructures}.

The output of the algorithm is dependent of the randomized nature of the initial neural map, sampled input data and the process parameters.
As a consequence, the resulting clusters differ for each SOM outcome.
The geometric sizes of resulting segmented cluster regions vary significantly  between independent realizations.

\subsection{Results of SCE segmentation}

In our showcase we stack a set of $62$ SOM cluster masks, which were obtained from the $15$ independent SOM runs with different process parameters. 
The goodness-of-fit metrics described in Section~\ref{atlastheory} were used to find most stable structures in the image and discard bad SOM maps.

\begin{figure*}
   \centering
    \centering\includegraphics[width=6.2cm]{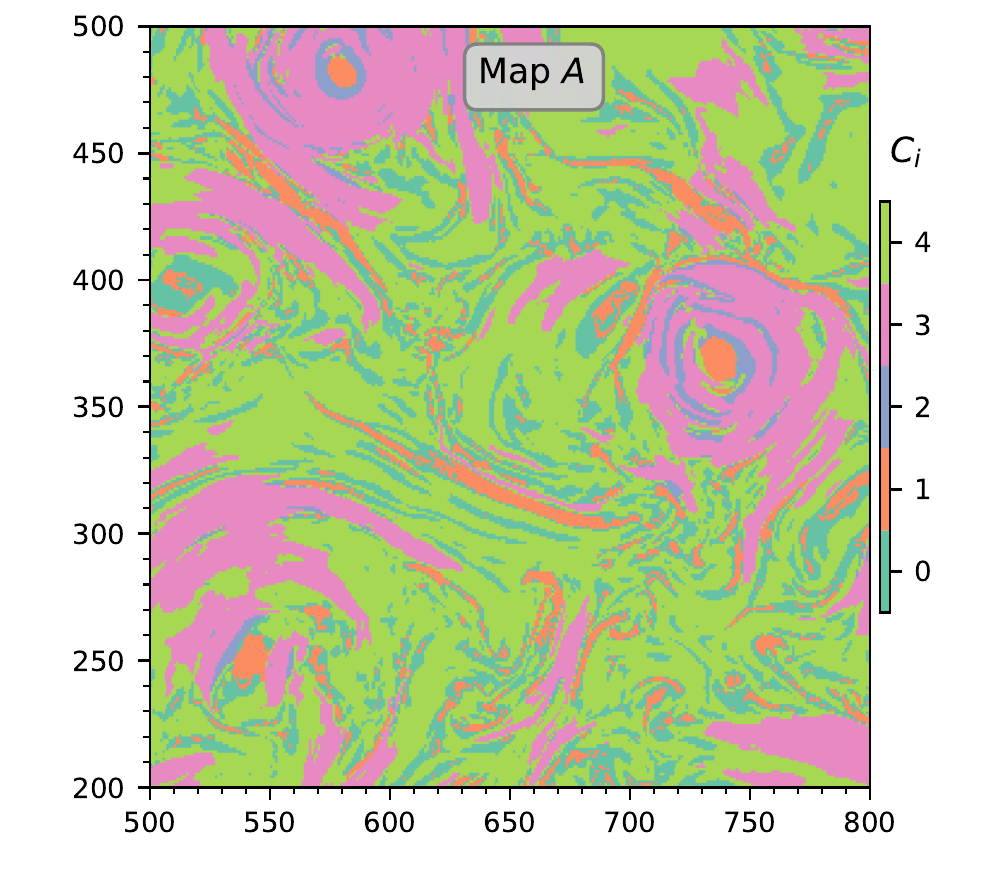}
    \centering\includegraphics[width=6.2cm]{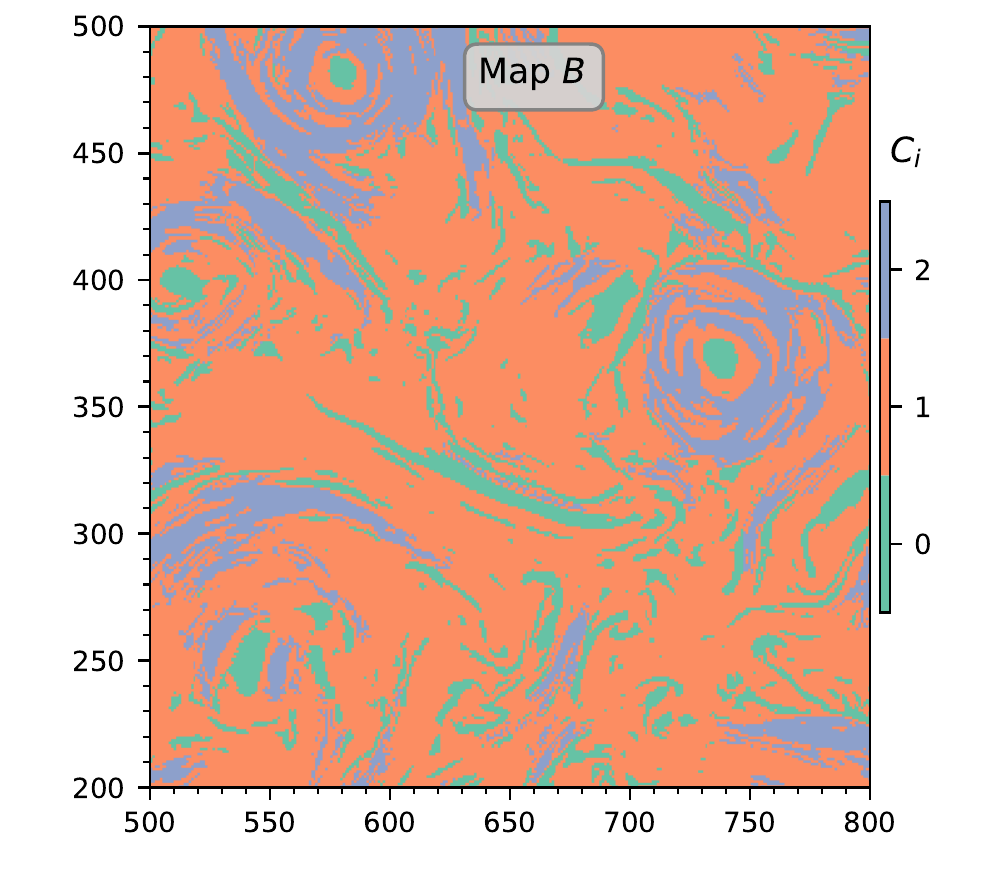}
    \centering\includegraphics[width=6.2cm]{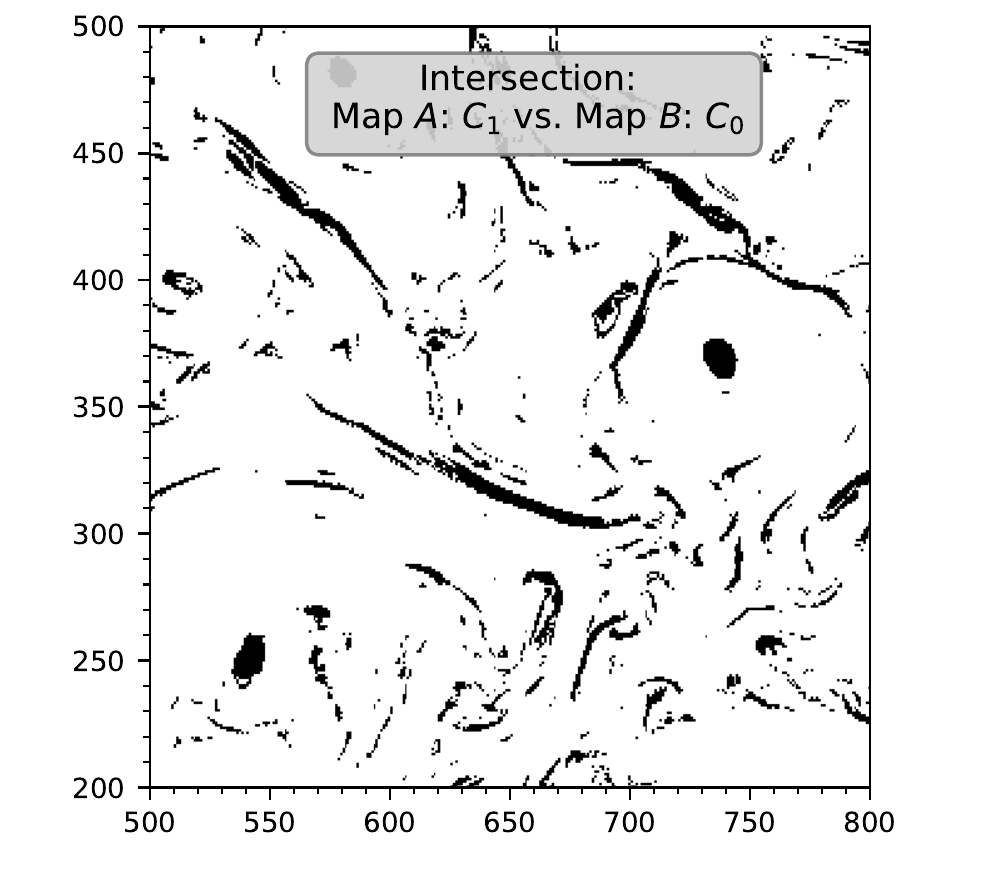}
    \centering\includegraphics[width=6.2cm]{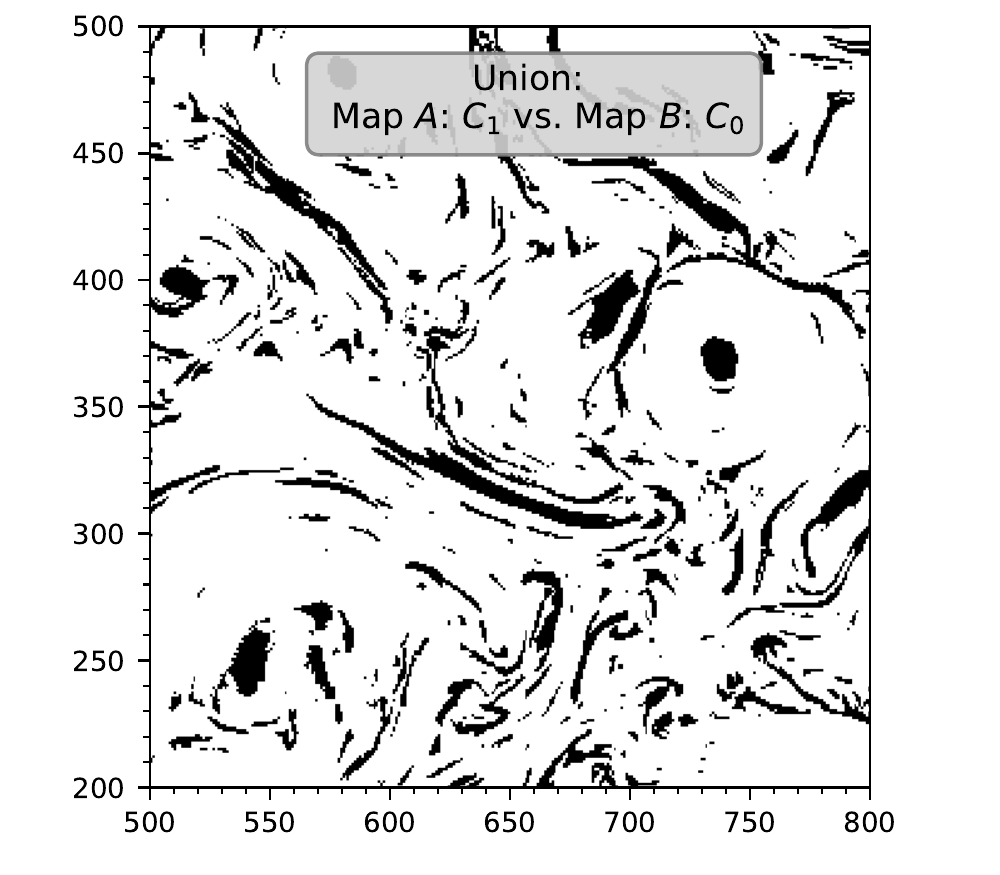}
  \caption{
      Visualization of the map-to-map operations performed between two SOM cluster map realizations.
      Top panels show a comparison of two SOM runs (focusing on a small region of the complete image) for which the pixel group information is projected back to the original image view. 
      The SOM map A has detected $5$ clusters and the SOM map B has detected $3$ clusters. 
      The lower-left panel shows the intersection matrix of $M_{1}^{A}$ ($C_1$ cluster mask of the SOM A) and $M_{0}^{B}$ ($C_0$ cluster mask of the SOM B). 
      The lower-right panel shows the union matrix for the same cluster mask combinations.
  } \label{IMG:Atlas1}
\end{figure*}

As an example, Figure~\ref{IMG:Atlas1} visualizes two cluster maps of two SOM algorithm results labeled as map $A$ and map $B$.
Map $A$ has detected $5$ clusters and map $B$ $3$ clusters. 
Thus, the corresponding mask sets are 
$\mathcal{M}^{A}=\{M_{0}^{A}$, $M_{1}^{A}$, $M_{2}^{A}$, $M_{3}^{A}$, $M_{4}^{A}\}$ and $\mathcal{M}^{B}=\{M_{0}^{B}$, $M_{1}^{B}$, $M_{2}^{B}\}$.
By taking the cluster mask $M_{1}^{A}$ from map $A$ and cluster mask $M_{0}^{B}$ from map $B$, we can calculate the intersection matrix $I$ (Equation~(\ref{EQ:intersection})) and union matrix $U$ (Equation~(\ref{EQ:union})).
Figure~\ref{IMG:Atlas1} bottom row images show the intersection $I$ and union $U$ matrices for these masks.
The pixels colored black correspond to the value $1$ and white to $0$. 
The $I$ matrix highlights the common pixels of $M_{1}^{A}$ 
and $M_{0}^{B}$. 
The $U$ matrix highlights pixels belonging to both $M_{1}^{A}$ and $M_{0}^{B}$. 
One can see that $I$ and $U$ are similarly shaped and have similarly positioned structures, suggesting that map $A$ and map $B$ have detected the same structure from the input image.

   \begin{figure*}
   \centering
       \centering\includegraphics[width=6.2cm]{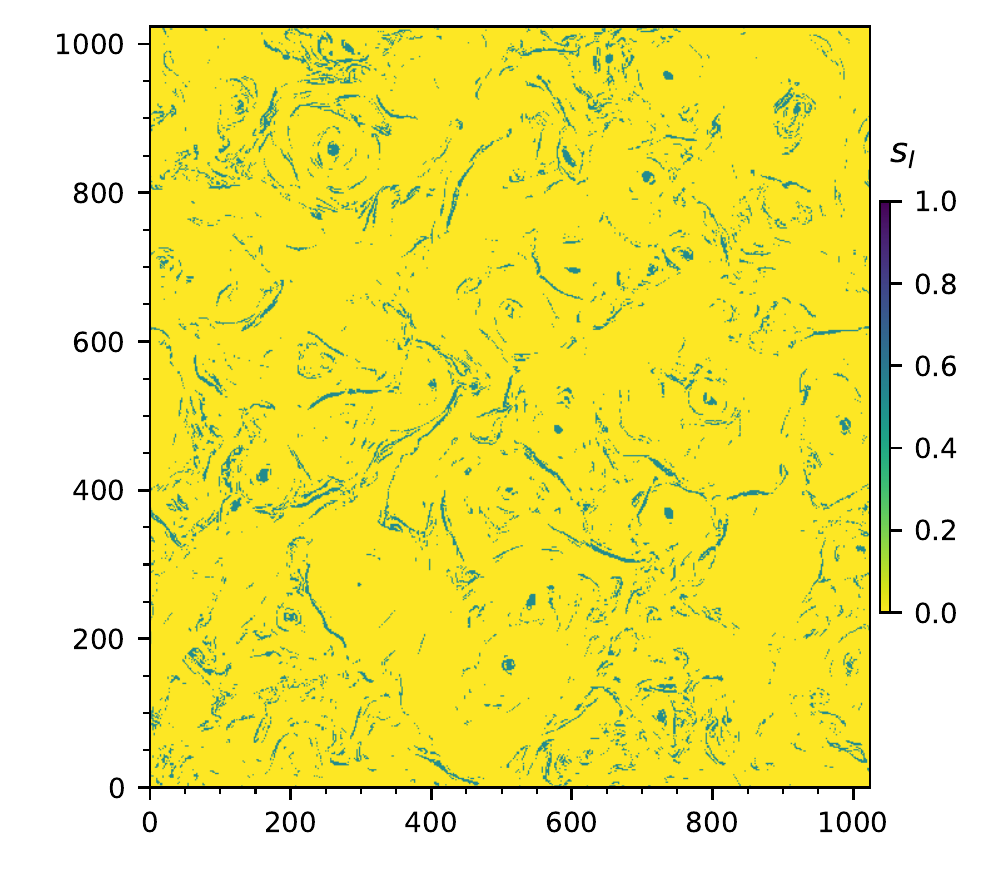}
    \centering\includegraphics[width=6.2cm]{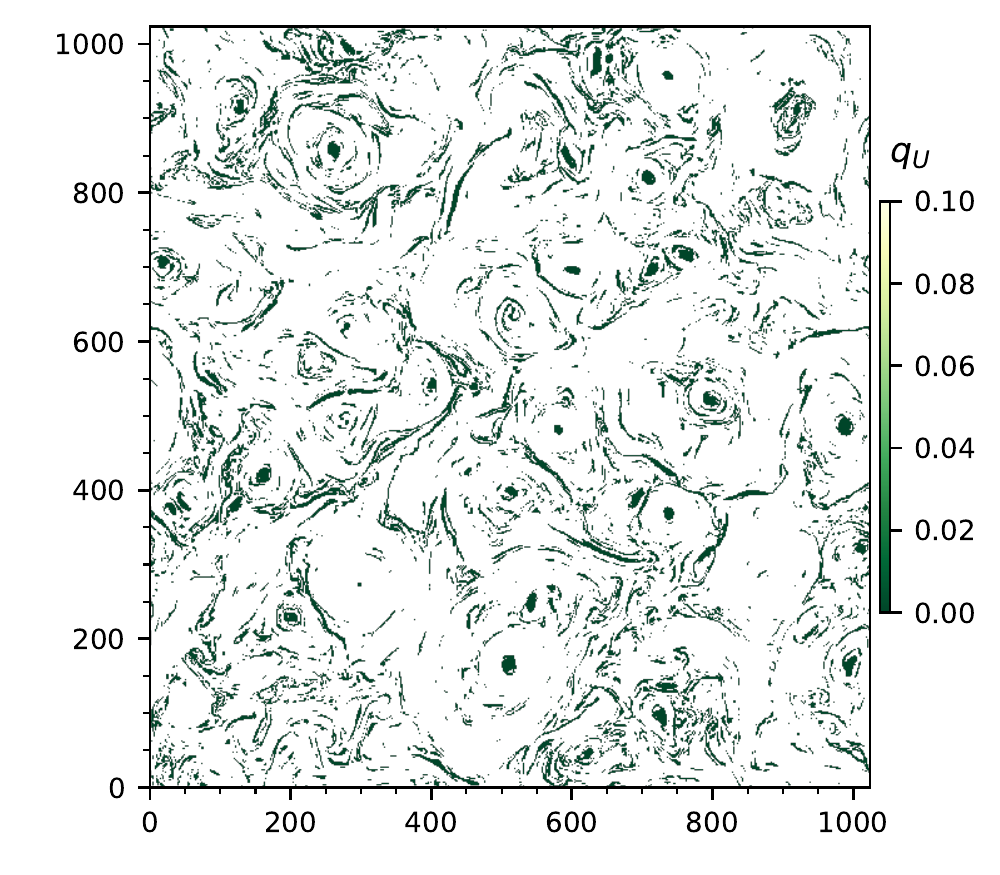}
  \caption{
  Resulting signal strength $s_I$ (left panel) and quality $q_U$ (right panel) values computed between the two clusters of map $A$ and map $B$ (visualized in Figure~\ref{IMG:Atlas1}).
  } \label{IMG: Atlas2}
\end{figure*}

The signal strength $s_{I}$ (Equation~(\ref{EQ:signalscalar})) for base mask $M_{1}^{A}$ and a mask $M_{0}^{B}$ is shown on the left panel of Figure~\ref{IMG: Atlas2}.
The metric compares the fraction of overlap between mask $M_{1}^{A}$ and mask $M_{0}^{B}$ (see Figure~\ref{IMG:Atlas1}).
Value of the quantity is the highest on the location where the two masks overlap.

The right panel of Figure~\ref{IMG: Atlas2} visualizes the quality of the union metric $q_{U}$ (Equation~(\ref{EQ:qualityscalar})). 
The $q_U$ depicts how well the two cluster masks $M_{1}^{A}$ and $M_{0}^{B}$ align on top of each other.
If their position is exactly the same, then the value $q_U$ for the pixels approaches $0$.
This corresponds to having identical masks.
In Figure ~\ref{IMG: Atlas2} the pixels with values close to $0$ indicate a good fit between the two masks.

\begin{figure}
   \centering
   \includegraphics[width=6.2cm]{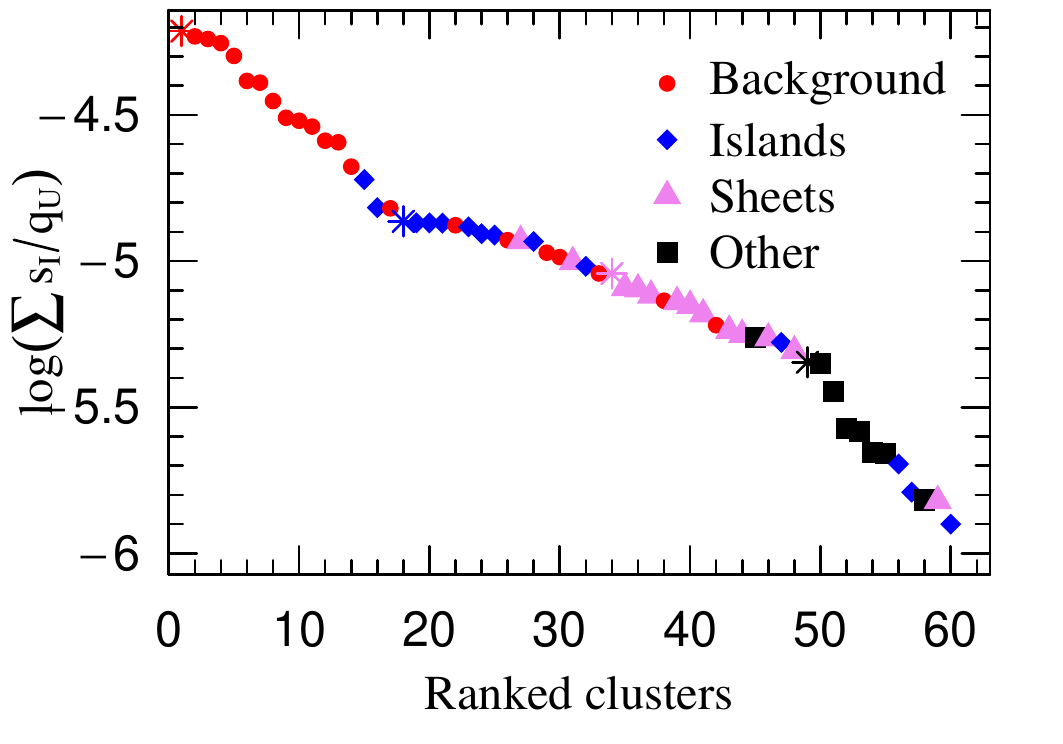}
     \caption{
    Total scalar values of the goodness-of-fit, 
    (see Equation~\ref{EQ:gsumscalar}) for all the $15$ independent SOM runs in the analysis.
    The clusters are shown in descending order.
    Each result was visually inspected and labeled into three different empirical categories:
    background pixels (red), circular islands (blue), sheets (violet), and other unclassified shapes (black). The cluster mask with the highest $g_{sum}$ value and the best accuracy inside each physical cluster is denoted with the star symbol.
    } \label{IMG: Atlas3}
\end{figure}
 
Figure~\ref{IMG: Atlas3} shows the total integrated scalars of goodness of fit, $g_{\mathrm{sum}}$ (Equation~(\ref{EQ:gsumscalar})) for all the masks detected by the $15$ SOM evaluations. 
We have visually inspected the resulting clusters and recovered the $3$ major physical structures from these results:
background pixels, 
circular islands, and 
sheets.

The background plasma (red points in Figure~\ref{IMG: Atlas3}) is detected the best among all the independent SOM algorithm evaluations, since the sum of the stacked $\frac{s_{I}}{q_{U}}$ values are the highest.
The second best cluster detected is that of the magnetic flux tubes or islands, (dark-blue diamonds in Figure~\ref{IMG: Atlas3}).
The third best detected structure is the current sheets (violet triangles in Figure~\ref{IMG: Atlas3}).
This manual classification is seen to correlate fairly well with the corresponding goodness-of-fit value.
A large change in the value of $g_{\mathrm{sum}}$ is seen to match well with the change of the physical meaning of the clusters.
As was discussed in Section~\ref{SEC:multiplemasks}, this fact could be further used to group different clusters between different SOM realizations together, since a same structure is expected to have a similar $g_{\mathrm{sum}}$ even if a different base map is used. 
Additionally, highly correlated base masks detecting same physical structures indicate a high accuracy for detecting the structure.

We also see that after the cluster ranked $50$th there is a sharp drop in the value of $g_{\mathrm{sum}}$ indicating that these clusters are only weakly correlated with other clusters in the SCE.
We use this fact to discard these points and select only the masks with rank $<50$ as having a meaningfully strong signal (in comparison to being just noise).
A statistically robust method to analyse, group, and select different data clusters based on their goodness-of-fit metrics is left for future work.

\begin{figure*}
   \centering
    \centering\includegraphics[width=6.2cm]{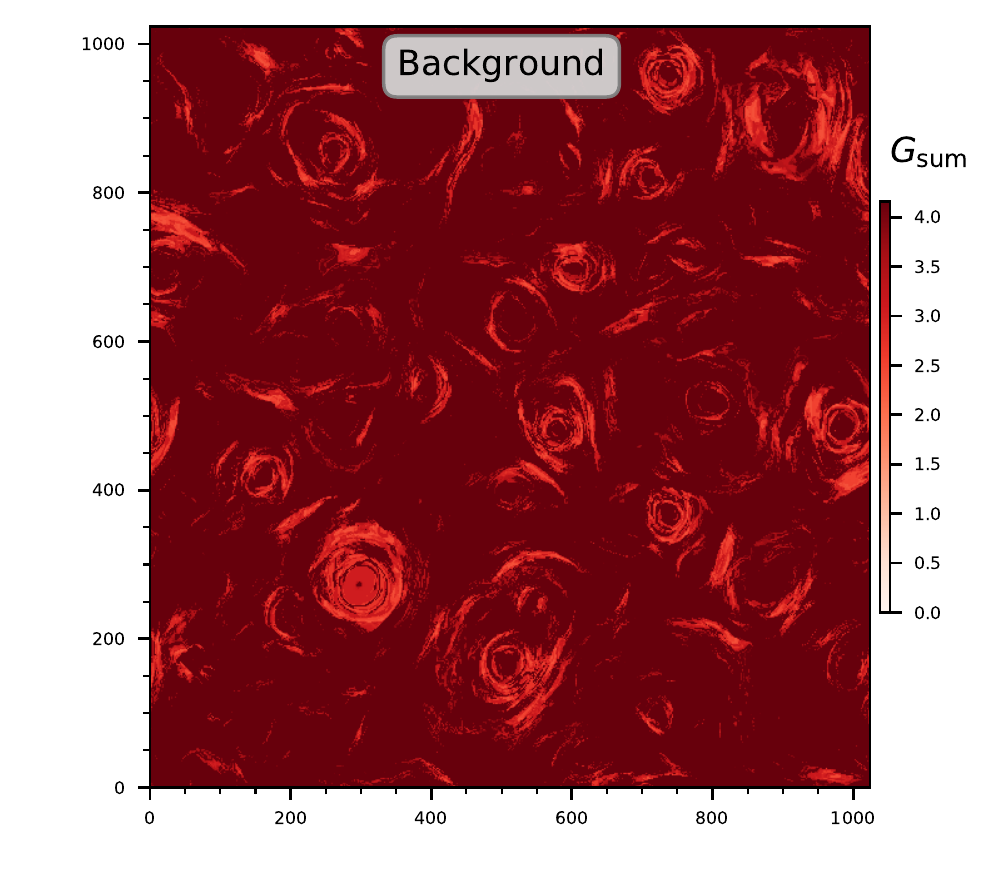}
    \centering\includegraphics[width=6.2cm]{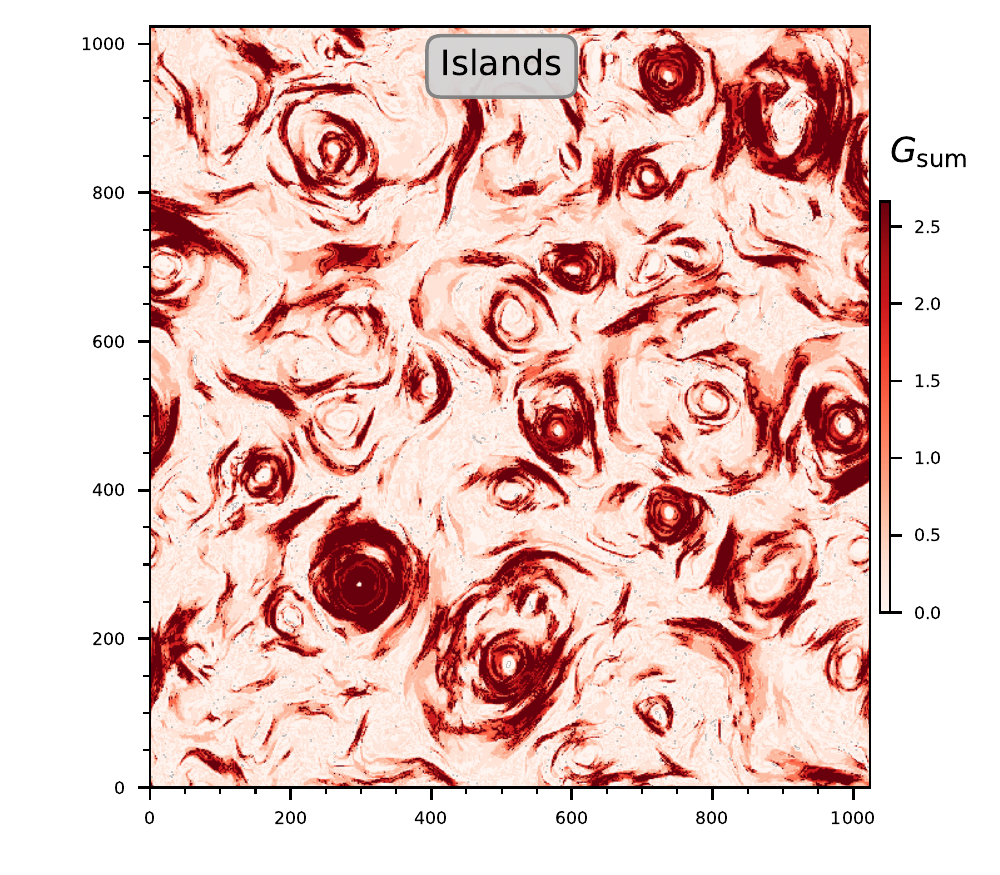}
    \centering\includegraphics[width=6.2cm]{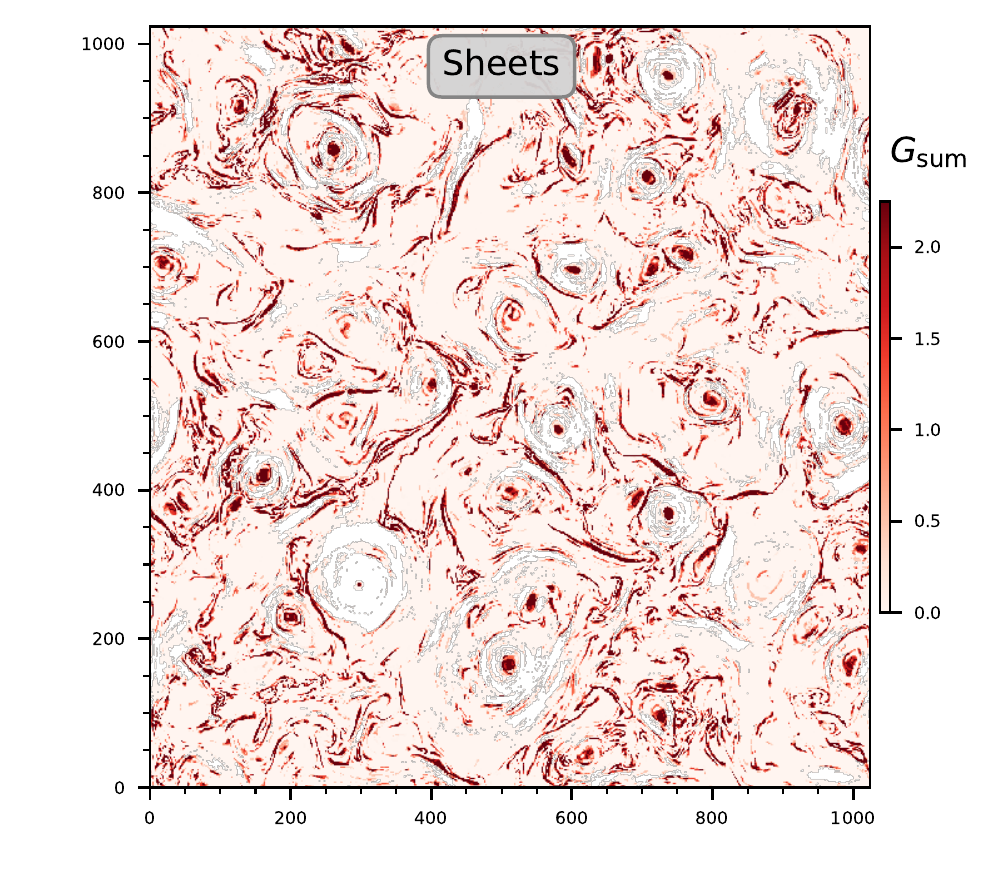}
    \centering\includegraphics[width=6.2cm]{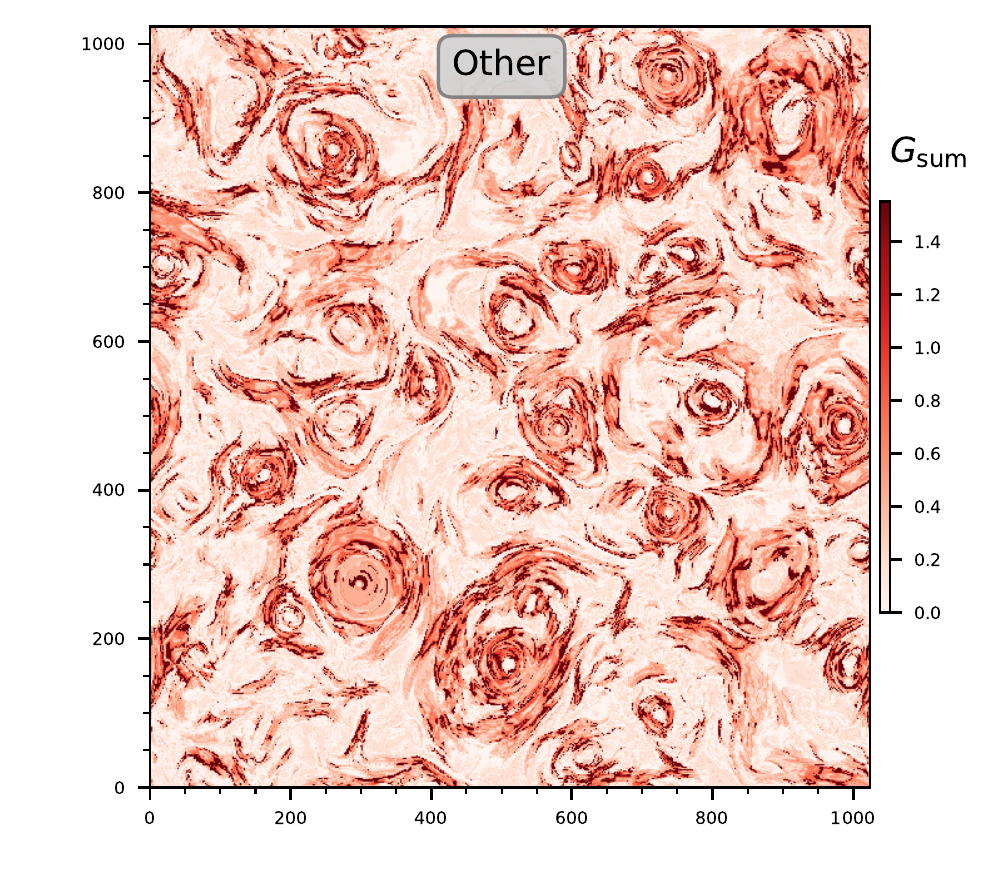}
  \caption{
  Goodness-of-fit matrices ($G_{\mathrm{sum}}$) of the four best detected SCE cluster maps (with $\log_{10}$-scale colors). 
  Each panel shows the highest ranking and highest accuracy cluster from Figure~\ref{IMG: Atlas3}, ordered based on their integrated goodness-of-fit quantity (Equation~\ref{EQ:gsumscalar}). 
  The three empirically derived categories are clearly visible in the results:
  background pixels (top-left panel),
  islands (top-right panel), and
  current sheets (bottom-left panel).
  The last bottom-right panel is seen to be a mixture of many clusters (mainly circular islands and sheets);
  it is correctly ranked as only the fourth best category and could therefore be safely discarded from any further analysis.} \label{IMG:Atlas4}
\end{figure*}

\subsection{Advantage of applying the SCE}

The best base mask clusters (highest $g_{\mathrm{sum}}$ values) with their corresponding goodness-of-fit matrix $G_{\mathrm{sum}}$ values are visualized in Figure~\ref{IMG:Atlas4}.
Here each pixel has accumulated a value (see Equation~(\ref{EQ:SQ_sum})) to the matrix $G_{\mathrm{sum}}$ from every comparison in the SCE.
These four panels describe the four best detected distinct clusters in the input data images. 
These $G_{\mathrm{sum}}$ matrices are the main outcome of the SCE algorithm and can be used at least in three different ways to help with the data analysis.
The four "winning" clusters of each physical cluster set are denote with star symbols in the Figure~\ref{IMG: Atlas3}.

Firstly, the activated pixels in each $G_{\mathrm{sum}}$ images are visually seen to capture each empirically defined classification category;
the benefit of the framework is that this work is now fully automated as it could also be performed by just ranking the masks by their $g_{\mathrm{sum}}$.
In the practical sense, these matrices can now be used for easily viewing the different physical structures that emerge from the clustering analysis.

Secondly, the resulting $G_{\mathrm{sum}}$ matrices provide a clear way of defining the actual segmentation boundaries:
we can now set a direct contour cutoff value for the $G_{\mathrm{sum}}$ providing a quantitative way of selecting which pixels belong to which cluster.
This is especially helpful since we need to perform statistically robust geometrical analysis of the resulting shapes.

Thirdly, throughout this analysis we have used the $8$ million pixels composed of a feature vector $X_{k}=(B_\perp$, $J_{\parallel}$, $[\vec{J}_\parallel \cdot\vec{E}])$.
However, nothing guarantees that these features are the best to detect these physical shapes.
Therefore, the presented framework can also be used to determine the best combination of the features for detecting the clusters of interest from the observed images.
In practice, the goodness of fit measures can be compared between the different SOM algorithm realizations that have been trained with distinct feature vectors on the same input data. 
The actual numerical value can be easily used to rank the different feature vector combinations.
We leave this analysis for further work.

\subsection{Analysis of SCE results}
The SCE framework was designed to combine decisions of independent unsupervised clustering results for image data. 
The clusters present in our astrophysical plasma simulations represent fine geometrical structures. 
The SCE algorithm is designed to tackle the prevailing issue of not gaining robust regions of interest for the geometrical structures of interest.
This is especially important for fine geometrical shapes, whose geometrical features such as width and length are of interest.

\begin{figure}
   \centering
   \includegraphics[width=8.5cm]{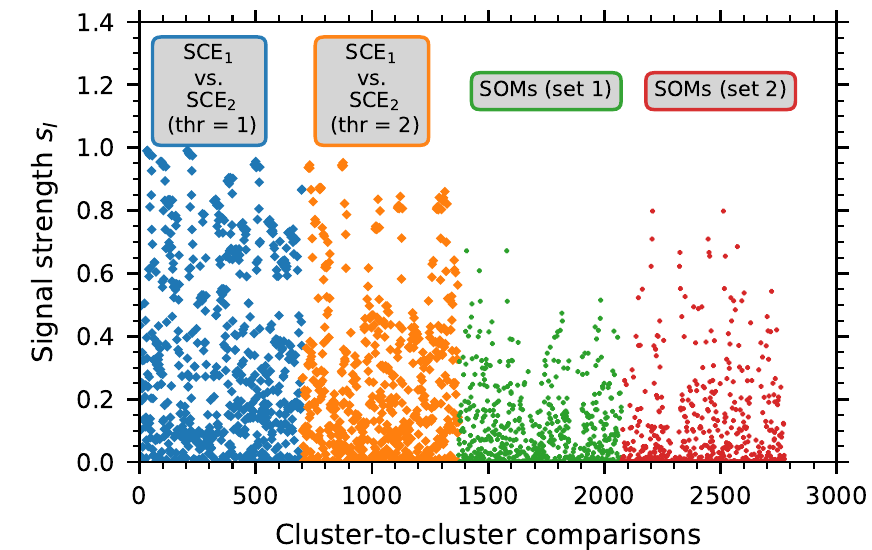}
     \caption{
    Pairwise comparison of cluster masks detected by two independent SCEs and the SOM sets that they are based on.
    Signal strength (equal to IOU in supervised learning) is used to quantitatively asses the similarity of the different masks.
    The 
    diamonds show pairwise comparisons of cluster masks between the two independent SCEs ($\mathrm{SCE}_1$ based on SOM set 1 and $\mathrm{SCE}_2$ based on SOM set 2);
    masks are generated using a threshold of $G_{\mathrm{sum}} > 1$ (blue) or $G_{\mathrm{sum}} > 2$ (orange).
    The points show comparison of clusters in SOM set 1 (green) and those in set 2 (red).
    Pairwise comparison of the cluster masks in the SOM sets are systematically lower than SCE results indicating that SCEs have more general masks in them.}
     \label{IMG: Analysis of SCE}
\end{figure}

Figure~\ref{IMG: Analysis of SCE} compares clusters from two independent ensembles, $\mathrm{SCE}_1$, and $\mathrm{SCE}_2$, and different SOM segmentations to asses the quantitative improvement of using the proposed method.
It shows pairwise comparisons of all clusters detected by the different methods. 
We use the signal strength as our comparison metric.
It is mathematically defined for the comparison of independent unsupervised clustering realizations in Section~\ref{Sect:signalstrength} (Equation~\ref{EQ:signalscalar}). The signal strength $s_I$ describes the amount of pixels similarly classified by the two classifiers in relation to the union of the compared clusters. 
The metric is equal to the IOU metric defined in supervised learning.
Each diamond on the Figure~\ref{IMG: Analysis of SCE}  represent the $s_I$ value between one cluster from $\mathrm{SCE}_1$ and another cluster from $\mathrm{SCE}_2$. 
The $s_I$ estimates the agreement between two independent cluster realizations. 
If the clusters consist of pixels describing the same phenomenon on the image, their $s_I$ will be close to 1. 
For the contrary, if the clusters describe completely different phenomenon, the metric will be close to 0. 

As described in Section~(\ref{SEC:multiplemasks}) the SCE framework actually results in a continuous $G_{\mathrm{sum}}$ scalar for each pixel on the image, and we therefore have a freedom in selecting how to quantize each mask.
We have tested that the SCE threshold level does not play an important role in the results by testing threshold values of $1$ and $2$ (blue vs. orange diamonds).

Figure ~\ref{IMG: Analysis of SCE} demonstrates that the SCE algorithm is capable of stabilizing the regions of interest for geometrical structures on images, the $s_I$ for $\mathrm{SCE_1}$ vs $\mathrm{SCE_2}$ is close to 1 for many clusters. 
The $s_I$ for the pairwise comparisons of independent clusters in SOM set 1 and in SOM set 2 are significantly lower and they reach the highest agreement of around 0.8.
We therefore conclude that the independent clusters from two different SCE results agree systematically better than the independent SOM cluster realizations.

We also note that the distributions of $s_I$ for all comparisons are actually bimodal, as expected theoretically.
The difference originates from comparing masks that have detected the same cluster (corresponding to high similarity and therefore high $s_I$) and from comparing physically different clusters (poor similarity and low $s_I$).
The cutoff value is present around $0.6$ for SCE comparisons and at $0.2$ for SOM comparisons.

\section{Conclusion}

We use computer vision and machine learning tools to automate the segmentation of
physical structures 
from 2-dimensional image snapshots originating from large supercomputer simulations of fully-kinetic turbulent plasma flows.
Machine learning clustering algorithms provide a fast and powerful method for detecting and segmenting visually distinct structures in data.
This makes them an attractive choice for automating data segmentation of computer simulation results:
Firstly, such algorithms are typically fast to evaluate which allows to couple them directly to the simulation time advance loops. 
Secondly, unsupervised algorithms need very little initial input from the user. 

Many such clustering algorithms are, however, non-deterministic and so the end result depends on the initial conditions and different technical process parameters that are used.
This can be an especially prohibiting feature in science applications where accurate and stable ROI boundaries are needed for e.g., in order to perform geometrical measurements of the segmented objects.
Here we designed a new machine learning framework that combines clustering results from multiple different segmentation realizations.
By averaging segmentation results from many independent clustering realizations the presented \textit{Statistically Combined Ensemble} (SCE) algorithm can yield much more accurate ROI objects.
The SCE algorithm can be used to

\begin{itemize}
\item determine optimal number of meaningful clusters pre\-sent in the input data,
\item uniquely identify each pixel with the corresponding object cluster, and
\item segment clustered pixels into stable contours that can be further analyzed for their geometrical shape, size, and area.
\end{itemize}
The SCE algorithm uses image cluster masks as a base unit for the stacking operations;
each individual clustering realization (obtained here via SOM algorithm) can be split into different image masks where only the pixels from one specific cluster are 'active'.
We defined stacking operations of these masks by using the union, intersection, and sum of two masks.
These quantities are shown to reflect the cluster signal strength and quality of the mask matching in each comparison.
The resulting quality and strength measures can be combined with a weighted average over the complete SCE.
This procedure enables to stack and combine information from many parallelly evaluated clustering realizations. 
This, in turn, enhances the accuracy of the detection algorithm and makes it suitable for use in many science applications where accurate and stable ROI boundaries are needed.

In the future, we plan to apply the SCE method for studying the spatial and temporal properties of intermittent turbulent structures found in the simulations discussed.

\section*{Acknowledgements}
We thank Pekka Manninen for useful discussions related to designing the stacking operations and the two anonymous referees for their comments and suggestions that helped to improve the paper.
MB would like to thank Elmo Tempel and Radu S. Stoica for support and discussions. 
MB acknowledges support from NORDITA via the Visiting PhD program and JN via the NORDITA Postdoctoral Fellowship.
MB acknowledges the financial support by the institutional research funding IUT40-2 of the Estonian Ministry of Education and Research and the support by the Centre of Excellence “Dark side of the Universe” (TK133), which is financed by the European Union through the European Regional Development Fund. 
The work of MB was also supported by the European Research Council Consolidator grant 682068-PRESTISSIMO.
The work has been partially performed under the Project HPC-EUROPA3 (INFRAIA- 2016-1-730897), with the support of the EC Research Innovation Action under the H2020 Programme.
The simulations were performed on resources provided by the Swedish National Infrastructure for Computing (SNIC) at PDC.

\bibliography{paper}

\begin{thebibliography}{42}
\expandafter\ifx\csname natexlab\endcsname\relax\def\natexlab#1{#1}\fi
\providecommand{\url}[1]{\texttt{#1}}
\providecommand{\href}[2]{#2}
\providecommand{\path}[1]{#1}
\providecommand{\DOIprefix}{doi:}
\providecommand{\ArXivprefix}{arXiv:}
\providecommand{\URLprefix}{URL: }
\providecommand{\Pubmedprefix}{pmid:}
\providecommand{\doi}[1]{\href{http://dx.doi.org/#1}{\path{#1}}}
\providecommand{\Pubmed}[1]{\href{pmid:#1}{\path{#1}}}
\providecommand{\bibinfo}[2]{#2}
\ifx\xfnm\relax \def\xfnm[#1]{\unskip,\space#1}\fi
\bibitem[{Biskamp(2003)}]{Biskamp_2003}
\bibinfo{author}{D.~Biskamp}, \bibinfo{title}{Magnetohydrodynamic Turbulence},
  \bibinfo{publisher}{Cambridge University Press}, \bibinfo{year}{2003}.
  \DOIprefix\doi{10.1017/cbo9780511535222}.
\bibitem[{{Uritsky} et~al.(2010){Uritsky}, {Pouquet}, {Rosenberg}, {Mininni},
  and {Donovan}}]{Uritsky2010}
\bibinfo{author}{V.~M. {Uritsky}}, \bibinfo{author}{A.~{Pouquet}},
  \bibinfo{author}{D.~{Rosenberg}}, \bibinfo{author}{P.~D. {Mininni}},
  \bibinfo{author}{E.~F. {Donovan}},
\newblock \bibinfo{title}{{Structures in magnetohydrodynamic turbulence:
  Detection and scaling}},
\newblock \bibinfo{journal}{Phys.~Rev.~E.} \bibinfo{volume}{82}
  (\bibinfo{year}{2010}) \bibinfo{pages}{056326}.
\bibitem[{{Zhdankin} et~al.(2013){Zhdankin}, {Uzdensky}, {Perez}, and
  {Boldyrev}}]{Zhdankin2013}
\bibinfo{author}{V.~{Zhdankin}}, \bibinfo{author}{D.~A. {Uzdensky}},
  \bibinfo{author}{J.~C. {Perez}}, \bibinfo{author}{S.~{Boldyrev}},
\newblock \bibinfo{title}{{Statistical Analysis of Current Sheets in
  Three-dimensional Magnetohydrodynamic Turbulence}},
\newblock \bibinfo{journal}{ApJ} \bibinfo{volume}{771} (\bibinfo{year}{2013})
  \bibinfo{pages}{124}.
\bibitem[{{Chatraee Azizabadi} et~al.(2020){Chatraee Azizabadi}, {Jain}, and
  {B{\"u}chner}}]{Azizabadi2020}
\bibinfo{author}{A.~{Chatraee Azizabadi}}, \bibinfo{author}{N.~{Jain}},
  \bibinfo{author}{J.~{B{\"u}chner}},
\newblock \bibinfo{title}{{Identification and characterization of current
  sheets in collisionless plasma turbulence}},
\newblock \bibinfo{journal}{arXiv e-prints}  (\bibinfo{year}{2020})
  \bibinfo{pages}{arXiv:2009.03881}.
\bibitem[{{Dupuis} et~al.(2020){Dupuis}, {Goldman}, {Newman}, {Amaya}, and
  {Lapenta}}]{2020Dupuis}
\bibinfo{author}{R.~{Dupuis}}, \bibinfo{author}{M.~V. {Goldman}},
  \bibinfo{author}{D.~L. {Newman}}, \bibinfo{author}{J.~{Amaya}},
  \bibinfo{author}{G.~{Lapenta}},
\newblock \bibinfo{title}{{Characterizing Magnetic Reconnection Regions Using
  Gaussian Mixture Models on Particle Velocity Distributions}},
\newblock \bibinfo{journal}{\apj} \bibinfo{volume}{889} (\bibinfo{year}{2020})
  \bibinfo{pages}{22}.
\bibitem[{{Hu} et~al.(2020){Hu}, {Sisti}, {Finelli}, {Califano}, {Dargent},
  {Faganello}, {Camporeale}, and {Teunissen}}]{Hu2020}
\bibinfo{author}{A.~{Hu}}, \bibinfo{author}{M.~{Sisti}},
  \bibinfo{author}{F.~{Finelli}}, \bibinfo{author}{F.~{Califano}},
  \bibinfo{author}{J.~{Dargent}}, \bibinfo{author}{M.~{Faganello}},
  \bibinfo{author}{E.~{Camporeale}}, \bibinfo{author}{J.~{Teunissen}},
\newblock \bibinfo{title}{{Identifying Magnetic Reconnection in 2D Hybrid
  Vlasov Maxwell Simulations with Convolutional Neural Networks}},
\newblock \bibinfo{journal}{ApJ} \bibinfo{volume}{900} (\bibinfo{year}{2020})
  \bibinfo{pages}{86}.
\bibitem[{{Sisti} et~al.(2021){Sisti}, {Finelli}, {Pedrazzi}, {Faganello},
  {Califano}, and {Delli Ponti}}]{2021Sisti}
\bibinfo{author}{M.~{Sisti}}, \bibinfo{author}{F.~{Finelli}},
  \bibinfo{author}{G.~{Pedrazzi}}, \bibinfo{author}{M.~{Faganello}},
  \bibinfo{author}{F.~{Califano}}, \bibinfo{author}{F.~{Delli Ponti}},
\newblock \bibinfo{title}{{Detecting Reconnection Events in Kinetic Vlasov
  Hybrid Simulations Using Clustering Techniques}},
\newblock \bibinfo{journal}{\apj} \bibinfo{volume}{908} (\bibinfo{year}{2021})
  \bibinfo{pages}{107}.
\bibitem[{He et~al.(2017)He, Gkioxari, Dollár, and Girshick}]{he2017mask}
\bibinfo{author}{K.~He}, \bibinfo{author}{G.~Gkioxari},
  \bibinfo{author}{P.~Dollár}, \bibinfo{author}{R.~Girshick},
  \bibinfo{title}{Mask r-cnn}, \bibinfo{year}{2017}.
  \href{http://arxiv.org/abs/1703.06870}{\tt arXiv:1703.06870}.
\bibitem[{Choy et~al.(2019)Choy, Gwak, and Savarese}]{choy20194d}
\bibinfo{author}{C.~Choy}, \bibinfo{author}{J.~Gwak},
  \bibinfo{author}{S.~Savarese}, \bibinfo{title}{4d spatio-temporal convnets:
  Minkowski convolutional neural networks}, \bibinfo{year}{2019}.
  \href{http://arxiv.org/abs/1904.08755}{\tt arXiv:1904.08755}.
\bibitem[{Valada et~al.(2019)Valada, Mohan, and Burgard}]{Valada_2019}
\bibinfo{author}{A.~Valada}, \bibinfo{author}{R.~Mohan},
  \bibinfo{author}{W.~Burgard},
\newblock \bibinfo{title}{Self-supervised model adaptation for multimodal
  semantic segmentation},
\newblock \bibinfo{journal}{International Journal of Computer Vision}
  \bibinfo{volume}{128} (\bibinfo{year}{2019}) \bibinfo{pages}{1239–1285}.
\bibitem[{{Landrieu} and {Simonovsky}(2018)}]{Landrieu2018PointCloud}
\bibinfo{author}{L.~{Landrieu}}, \bibinfo{author}{M.~{Simonovsky}},
\newblock \bibinfo{title}{Large-scale point cloud semantic segmentation with
  superpoint graphs},
\newblock in: \bibinfo{booktitle}{2018 IEEE/CVF Conference on Computer Vision
  and Pattern Recognition}, \bibinfo{year}{2018}, pp.
  \bibinfo{pages}{4558--4567}.
\bibitem[{Wang et~al.(2017)Wang, Girshick, Gupta, and He}]{wang2017nonlocal}
\bibinfo{author}{X.~Wang}, \bibinfo{author}{R.~Girshick},
  \bibinfo{author}{A.~Gupta}, \bibinfo{author}{K.~He},
  \bibinfo{title}{Non-local neural networks}, \bibinfo{year}{2017}.
  \href{http://arxiv.org/abs/1711.07971}{\tt arXiv:1711.07971}.
\bibitem[{Zhu et~al.(2019)Zhu, Xu, Bai, Huang, and Bai}]{Zhu2019AsymmetricNN}
\bibinfo{author}{Z.~Zhu}, \bibinfo{author}{M.~Xu}, \bibinfo{author}{S.~Bai},
  \bibinfo{author}{T.~Huang}, \bibinfo{author}{X.~Bai},
\newblock \bibinfo{title}{Asymmetric non-local neural networks for semantic
  segmentation},
\newblock \bibinfo{journal}{2019 IEEE/CVF International Conference on Computer
  Vision (ICCV)}  (\bibinfo{year}{2019}) \bibinfo{pages}{593--602}.
\bibitem[{Hurtado and Alvarez(2001)}]{HURTADO2001}
\bibinfo{author}{J.~E. Hurtado}, \bibinfo{author}{D.~A. Alvarez},
\newblock \bibinfo{title}{Neural-network-based reliability analysis: a
  comparative study},
\newblock \bibinfo{journal}{Computer Methods in Applied Mechanics and
  Engineering} \bibinfo{volume}{191} (\bibinfo{year}{2001}) \bibinfo{pages}{113
  -- 132}. \bibinfo{note}{Micromechanics of Brittle Materials and Stochastic
  Analysis of Mechanical Systems}.
\bibitem[{Shrestha and Mahmood(2019)}]{Shrestha2019}
\bibinfo{author}{A.~Shrestha}, \bibinfo{author}{A.~Mahmood},
\newblock \bibinfo{title}{Review of deep learning algorithms and
  architectures},
\newblock \bibinfo{journal}{IEEE Access} \bibinfo{volume}{PP}
  (\bibinfo{year}{2019}) \bibinfo{pages}{1--1}.
\bibitem[{Gansbeke et~al.(2020)Gansbeke, Vandenhende, Georgoulis, Proesmans,
  and Gool}]{gansbeke2020scan}
\bibinfo{author}{W.~V. Gansbeke}, \bibinfo{author}{S.~Vandenhende},
  \bibinfo{author}{S.~Georgoulis}, \bibinfo{author}{M.~Proesmans},
  \bibinfo{author}{L.~V. Gool}, \bibinfo{title}{Scan: Learning to classify
  images without labels}, \bibinfo{year}{2020}.
  \href{http://arxiv.org/abs/2005.12320}{\tt arXiv:2005.12320}.
\bibitem[{Ventura et~al.(2019)Ventura, Bellver, Girbau, Salvador, Marqu{\'e}s,
  and i~Nieto}]{Ventura2019RVOSER}
\bibinfo{author}{C.~Ventura}, \bibinfo{author}{M.~Bellver},
  \bibinfo{author}{A.~Girbau}, \bibinfo{author}{A.~Salvador},
  \bibinfo{author}{F.~Marqu{\'e}s}, \bibinfo{author}{X.~G. i~Nieto},
\newblock \bibinfo{title}{Rvos: End-to-end recurrent network for video object
  segmentation},
\newblock \bibinfo{journal}{2019 IEEE/CVF Conference on Computer Vision and
  Pattern Recognition (CVPR)}  (\bibinfo{year}{2019})
  \bibinfo{pages}{5272--5281}.
\bibitem[{Halkidi et~al.(2001)Halkidi, Batistakis, and
  Vazirgiannis}]{halkidi2001clustering}
\bibinfo{author}{M.~Halkidi}, \bibinfo{author}{Y.~Batistakis},
  \bibinfo{author}{M.~Vazirgiannis},
\newblock \bibinfo{title}{On clustering validation techniques},
\newblock \bibinfo{journal}{Journal of intelligent information systems}
  \bibinfo{volume}{17} (\bibinfo{year}{2001}) \bibinfo{pages}{107--145}.
\bibitem[{Rini et~al.(2018)Rini, Novianti, and Fransiska}]{Rini_2018}
\bibinfo{author}{D.~S. Rini}, \bibinfo{author}{P.~Novianti},
  \bibinfo{author}{H.~Fransiska},
\newblock \bibinfo{title}{Internal cluster validation on earthquake data in the
  province of bengkulu},
\newblock \bibinfo{journal}{{IOP} Conference Series: Materials Science and
  Engineering} \bibinfo{volume}{335} (\bibinfo{year}{2018})
  \bibinfo{pages}{012048}.
\bibitem[{{Fred} and {Jain}(2002)}]{Fred2002}
\bibinfo{author}{A.~L.~N. {Fred}}, \bibinfo{author}{A.~K. {Jain}},
\newblock \bibinfo{title}{Data clustering using evidence accumulation},
\newblock in: \bibinfo{booktitle}{Object recognition supported by user
  interaction for service robots}, volume~\bibinfo{volume}{4},
  \bibinfo{year}{2002}, pp. \bibinfo{pages}{276--280 vol.4}.
  \DOIprefix\doi{10.1109/ICPR.2002.1047450}.
\bibitem[{Zhang et~al.(2014)Zhang, Yang, Jia, Kasabov, Jia, and
  Zhou}]{zhang2014unsupervised}
\bibinfo{author}{W.~Zhang}, \bibinfo{author}{J.~Yang},
  \bibinfo{author}{W.~Jia}, \bibinfo{author}{N.~Kasabov},
  \bibinfo{author}{Z.~Jia}, \bibinfo{author}{L.~Zhou},
\newblock \bibinfo{title}{Unsupervised segmentation using cluster ensembles},
\newblock in: \bibinfo{booktitle}{International Conference on Neural
  Information Processing}, \bibinfo{organization}{Springer},
  \bibinfo{year}{2014}, pp. \bibinfo{pages}{76--84}.
\bibitem[{Topchy et~al.(2004)Topchy, Jain, and Punch}]{topchy2004mixture}
\bibinfo{author}{A.~Topchy}, \bibinfo{author}{A.~K. Jain},
  \bibinfo{author}{W.~Punch},
\newblock \bibinfo{title}{A mixture model for clustering ensembles},
\newblock in: \bibinfo{booktitle}{Proceedings of the 2004 SIAM international
  conference on data mining}, \bibinfo{organization}{SIAM},
  \bibinfo{year}{2004}, pp. \bibinfo{pages}{379--390}.
\bibitem[{Fern and Brodley(2004)}]{GraphPartitioning1}
\bibinfo{author}{X.~Z. Fern}, \bibinfo{author}{C.~E. Brodley},
\newblock \bibinfo{title}{Solving cluster ensemble problems by bipartite graph
  partitioning},
\newblock in: \bibinfo{booktitle}{Proceedings of the Twenty-First International
  Conference on Machine Learning}, ICML '04, \bibinfo{publisher}{Association
  for Computing Machinery}, \bibinfo{address}{New York, NY, USA},
  \bibinfo{year}{2004}, p.~\bibinfo{pages}{36}. \URLprefix
  \url{https://doi.org/10.1145/1015330.1015414}.
  \DOIprefix\doi{10.1145/1015330.1015414}.
\bibitem[{Strehl and Ghosh(2002)}]{strehl2002cluster}
\bibinfo{author}{A.~Strehl}, \bibinfo{author}{J.~Ghosh},
\newblock \bibinfo{title}{Cluster ensembles---a knowledge reuse framework for
  combining multiple partitions},
\newblock \bibinfo{journal}{Journal of machine learning research}
  \bibinfo{volume}{3} (\bibinfo{year}{2002}) \bibinfo{pages}{583--617}.
\bibitem[{Yu et~al.(2014{\natexlab{a}})Yu, Rui, and Chen}]{YuIEEE}
\bibinfo{author}{J.~Yu}, \bibinfo{author}{Y.~Rui}, \bibinfo{author}{B.~Chen},
\newblock \bibinfo{title}{Exploiting click constraints and multi-view features
  for image re-ranking},
\newblock \bibinfo{journal}{IEEE Transactions on Multimedia}
  \bibinfo{volume}{16} (\bibinfo{year}{2014}{\natexlab{a}})
  \bibinfo{pages}{159--168}.
\bibitem[{Yu et~al.(2014{\natexlab{b}})Yu, Rui, and Tao}]{YuIEEE2}
\bibinfo{author}{J.~Yu}, \bibinfo{author}{Y.~Rui}, \bibinfo{author}{D.~Tao},
\newblock \bibinfo{title}{Click prediction for web image reranking using
  multimodal sparse coding},
\newblock \bibinfo{journal}{IEEE Transactions on Image Processing}
  \bibinfo{volume}{23} (\bibinfo{year}{2014}{\natexlab{b}})
  \bibinfo{pages}{2019--2032}.
\bibitem[{Jiang and Zhou(2004)}]{jiang2004som}
\bibinfo{author}{Y.~Jiang}, \bibinfo{author}{Z.-H. Zhou},
\newblock \bibinfo{title}{Som ensemble-based image segmentation},
\newblock \bibinfo{journal}{Neural Processing Letters} \bibinfo{volume}{20}
  (\bibinfo{year}{2004}) \bibinfo{pages}{171--178}.
\bibitem[{{Kohonen}(2001)}]{kohonenSOM}
\bibinfo{author}{T.~{Kohonen}}, \bibinfo{title}{Self-Organizing maps},
  \bibinfo{publisher}{Springer-Verlag Berlin Heidelberg}, \bibinfo{year}{2001}.
\bibitem[{Kohonen(2013)}]{kohonen2013essentials}
\bibinfo{author}{T.~Kohonen},
\newblock \bibinfo{title}{Essentials of the self-organizing map},
\newblock \bibinfo{journal}{Neural networks} \bibinfo{volume}{37}
  (\bibinfo{year}{2013}) \bibinfo{pages}{52--65}.
\bibitem[{{Qiu} et~al.(2020){Qiu}, {Gamba}, {Schmitt}, and {Zhu}}]{2020Qiu}
\bibinfo{author}{C.~{Qiu}}, \bibinfo{author}{P.~{Gamba}},
  \bibinfo{author}{M.~{Schmitt}}, \bibinfo{author}{X.~X. {Zhu}},
\newblock \bibinfo{title}{{Learning from Noisy Samples for Man-Made Impervious
  Surface Mapping}},
\newblock \bibinfo{journal}{ISPRS Annals of Photogrammetry, Remote Sensing and
  Spatial Information Sciences} \bibinfo{volume}{5.3} (\bibinfo{year}{2020})
  \bibinfo{pages}{787--794}.
\bibitem[{{Yu} et~al.(2020){Yu}, {Pang}, {Xu}, and {Liang}}]{2020Xiangchun}
\bibinfo{author}{X.~{Yu}}, \bibinfo{author}{W.~{Pang}},
  \bibinfo{author}{Q.~{Xu}}, \bibinfo{author}{M.~{Liang}},
\newblock \bibinfo{title}{{Mammographic image classification with deep fusion
  learning}},
\newblock \bibinfo{journal}{Scientific Reports} \bibinfo{volume}{10}
  (\bibinfo{year}{2020}) \bibinfo{pages}{14361}.
\bibitem[{{Zhdankin} et~al.(2017){Zhdankin}, {Werner}, {Uzdensky}, and
  {Begelman}}]{Zhdankin2017}
\bibinfo{author}{V.~{Zhdankin}}, \bibinfo{author}{G.~R. {Werner}},
  \bibinfo{author}{D.~A. {Uzdensky}}, \bibinfo{author}{M.~C. {Begelman}},
\newblock \bibinfo{title}{{Kinetic Turbulence in Relativistic Plasma: From
  Thermal Bath to Nonthermal Continuum}},
\newblock \bibinfo{journal}{Phys.~Rev.~Lett.} \bibinfo{volume}{118}
  (\bibinfo{year}{2017}) \bibinfo{pages}{055103}.
\bibitem[{{Comisso} and {Sironi}(2018)}]{Comisso2018}
\bibinfo{author}{L.~{Comisso}}, \bibinfo{author}{L.~{Sironi}},
\newblock \bibinfo{title}{{Particle Acceleration in Relativistic Plasma
  Turbulence}},
\newblock \bibinfo{journal}{Phys.~Rev.~Lett.} \bibinfo{volume}{121}
  (\bibinfo{year}{2018}) \bibinfo{pages}{255101}.
\bibitem[{{N{\"a}ttil{\"a}}(2019)}]{Nattila2019}
\bibinfo{author}{J.~{N{\"a}ttil{\"a}}},
\newblock \bibinfo{title}{{Runko: Modern multi-physics toolbox for simulating
  plasma}},
\newblock \bibinfo{journal}{arXiv e-prints}  (\bibinfo{year}{2019})
  \bibinfo{pages}{arXiv:1906.06306}.
\bibitem[{{Kohonen} and {M{\"a}kisara}(1989)}]{Kohonen1989}
\bibinfo{author}{T.~{Kohonen}}, \bibinfo{author}{K.~{M{\"a}kisara}},
\newblock \bibinfo{title}{{The self-organizing feature maps}},
\newblock \bibinfo{journal}{Phys.~Scr} \bibinfo{volume}{39}
  (\bibinfo{year}{1989}) \bibinfo{pages}{168--172}.
\bibitem[{Valova et~al.(2013)Valova, Georgiev, Gueorguieva, and
  Olson}]{valova2013initialization}
\bibinfo{author}{I.~Valova}, \bibinfo{author}{G.~Georgiev},
  \bibinfo{author}{N.~Gueorguieva}, \bibinfo{author}{J.~Olson},
\newblock \bibinfo{title}{Initialization issues in self-organizing maps},
\newblock \bibinfo{journal}{Procedia Computer Science} \bibinfo{volume}{20}
  (\bibinfo{year}{2013}) \bibinfo{pages}{52--57}.
\bibitem[{Lee and Verleysen(2002)}]{lee2002self}
\bibinfo{author}{J.~A. Lee}, \bibinfo{author}{M.~Verleysen},
\newblock \bibinfo{title}{Self-organizing maps with recursive neighborhood
  adaptation},
\newblock \bibinfo{journal}{Neural networks} \bibinfo{volume}{15}
  (\bibinfo{year}{2002}) \bibinfo{pages}{993--1003}.
\bibitem[{Stefanovi{\v{c}} and Kurasova(2011)}]{stefanovivc2011influence}
\bibinfo{author}{P.~Stefanovi{\v{c}}}, \bibinfo{author}{O.~Kurasova},
\newblock \bibinfo{title}{Influence of learning rates and neighboring functions
  on self-organizing maps},
\newblock in: \bibinfo{booktitle}{International Workshop on Self-Organizing
  Maps}, \bibinfo{organization}{Springer}, \bibinfo{year}{2011}, pp.
  \bibinfo{pages}{141--150}.
\bibitem[{{De Bodt} et~al.(2007){De Bodt}, {Cottrell}, and
  {Verleysen}}]{statisticsSOM}
\bibinfo{author}{E.~{De Bodt}}, \bibinfo{author}{M.~{Cottrell}},
  \bibinfo{author}{M.~{Verleysen}},
\newblock \bibinfo{title}{{Statistical tools to assess the reliability of
  self-organizing maps}},
\newblock \bibinfo{journal}{arXiv Mathematics e-prints}  (\bibinfo{year}{2007})
  \bibinfo{pages}{math/0701144}.
\bibitem[{Jaffe et~al.(2016)Jaffe, Fetaya, Nadler, Jiang, and
  Kluger}]{pmlr-v51-jaffe16}
\bibinfo{author}{A.~Jaffe}, \bibinfo{author}{E.~Fetaya},
  \bibinfo{author}{B.~Nadler}, \bibinfo{author}{T.~Jiang},
  \bibinfo{author}{Y.~Kluger},
\newblock \bibinfo{title}{Unsupervised ensemble learning with dependent
  classifiers},
\newblock volume~\bibinfo{volume}{51} of \textit{\bibinfo{series}{Proceedings
  of Machine Learning Research}}, \bibinfo{publisher}{PMLR},
  \bibinfo{address}{Cadiz, Spain}, \bibinfo{year}{2016}, pp.
  \bibinfo{pages}{351--360}. \URLprefix
  \url{http://proceedings.mlr.press/v51/jaffe16.html}.
\bibitem[{Platanios et~al.(2017)Platanios, Poon, Mitchell, and
  Horvitz}]{Platanios2017}
\bibinfo{author}{E.~A. Platanios}, \bibinfo{author}{H.~Poon},
  \bibinfo{author}{T.~M. Mitchell}, \bibinfo{author}{E.~Horvitz},
\newblock \bibinfo{title}{Estimating accuracy from unlabeled data: A
  probabilistic logic approach},
\newblock in: \bibinfo{booktitle}{Proceedings of the 31st International
  Conference on Neural Information Processing Systems}, NIPS'17,
  \bibinfo{publisher}{Curran Associates Inc.}, \bibinfo{address}{Red Hook, NY,
  USA}, \bibinfo{year}{2017}, p. \bibinfo{pages}{4364–4373}.
\bibitem[{Rokach(2009)}]{ROKACH2009}
\bibinfo{author}{L.~Rokach},
\newblock \bibinfo{title}{Collective-agreement-based pruning of ensembles},
\newblock \bibinfo{journal}{Computational Statistics \& Data Analysis}
  \bibinfo{volume}{53} (\bibinfo{year}{2009}) \bibinfo{pages}{1015 -- 1026}.

\end{thebibliography}

\end{document}